\documentclass[lettersize,journal]{IEEEtran}
\usepackage{amsmath,amsfonts,amssymb,amsthm}
\usepackage{algorithm}
\usepackage{algpseudocode}
\usepackage{array}
\usepackage[caption=false,font=normalsize,labelfont=sf,textfont=sf]{subfig}
\usepackage{textcomp}
\usepackage{url}
\usepackage{verbatim}
\usepackage{graphicx}
\def\BibTeX{{\rm B\kern-.05em{\sc i\kern-.025em b}\kern-.08em
    T\kern-.1667em\lower.7ex\hbox{E}\kern-.125emX}}
\usepackage{balance}
\usepackage{hyperref}
\usepackage{xcolor}
\usepackage{multirow}
\usepackage{mathtools}
\usepackage{color,soul}

\DeclarePairedDelimiter{\ceil}{\lceil}{\rceil}

\begin{document}

\title{Airspace-aware Contingency Landing Planning}

\author{H. Emre Tekaslan and Ella M. Atkins
\thanks{This work was supported in part by NASA Langley Research Center RSES contract 80LARC23DA003.}
\thanks{H. Emre Tekaslan is with the Kevin T. Crofton Department of Aerospace and Ocean Engineering,
        Virginia Polytechnic Institute and State University, Blacksburg, VA 24060, USA,
        {Corresponding author: \tt\small tekaslan@vt.edu}}%
\thanks{Ella M. Atkins is with the Kevin T. Crofton Department of Aerospace and Ocean Engineering,
        Virginia Polytechnic Institute and State University,
        Blacksburg, VA 24060, USA,
        {\tt\small ematkins@vt.edu}}%
}


\maketitle

\begin{abstract}
This paper develops a real-time, search-based aircraft contingency landing planner that minimizes traffic disruptions while accounting for ground risk. The airspace model captures dense air traffic departure and arrival flows, helicopter corridors, and prohibited zones and is demonstrated with a Washington, D.C., area case study. Historical Automatic Dependent Surveillance–Broadcast (ADS-B) data are processed to estimate air traffic density. A low-latency computational geometry algorithm generates proximity-based heatmaps around high-risk corridors and restricted regions. Airspace risk is quantified as the cumulative exposure time of a landing trajectory within congested regions, while ground risk is assessed from overflown population density to jointly guide trajectory selection. A landing site selection module further mitigates disruption to nominal air traffic operations. Benchmarking against minimum-risk Dubins solutions demonstrates that the proposed planner achieves lower joint risk and reduced airspace disruption while maintaining real-time performance. Under airspace-risk-only conditions, the planner generates trajectories within an average of 2.9 seconds on a laptop computer. Future work will incorporate dynamic air traffic updates to enable spatiotemporal contingency landing planning that minimizes the need for real-time traffic rerouting.
\end{abstract}
\section{Introduction}
\label{sec:intro}
In-flight emergencies may require a distressed aircraft to initiate an immediate approach to landing—a response that can substantially disrupt air traffic flow. 
Departures may face delays, and arrivals may require rerouting, holding, or extended approaches to maintain separation. Although emergency aircraft receive operational priority, effective emergency landing is not merely a vehicle-level motion-planning problem; it is a system-level air transportation challenge unfolding within a congested airspace. Past events illustrate these vulnerabilities. For instance, US Airways Flight 1549 demonstrated how low altitude and dense terminal airspace can render voice-based coordination with delays insufficient \cite{Atkins2010}, while the Washington, D.C. Blackhawk–CRJ700 mid-air collision highlighted how interactions between rapidly evolving flight paths \cite{dc_crash} can lead to catastrophic outcomes even under normal operations. Together, these incidents underscore the limitations of human-only coordination in time-critical complex airspace conditions.

To strengthen resilience in these scenarios, the national airspace would benefit from an Assured Contingency Landing Management (ACLM) capability \cite{kim2025runtime} that provides real-time, feasible landing plans and coordinates with surrounding traffic through datalink \cite{ozmen2024survey}. This paper introduces such an airspace-aware emergency landing planning framework for wing-lift aircraft with degraded performance, explicitly modeling airspace complexity to improve safety by reducing exposure to high-density traffic corridors. When feasible, the planner produces a conflict-free trajectory that fits with ongoing operations, avoiding secondary conflicts and diversions; otherwise, it identifies and highlights unavoidable transitions across traffic corridors and minimizes transition time. By integrating airspace corridor and historic traffic flow data rather than focusing on aircraft-specific procedures, the framework supports scalable, airspace-aware contingency management.

Air conflict resolution has been a significant focus. Ref. \cite{osipychev2023} proposed a two-dimensional surrogate optimization method for obstacle avoidance and horizontal separation between two aircraft. Mixed-integer optimization techniques were applied to regulate speed \cite{mixed_integer} and heading \cite{Spencer_Trindade_2025} to ensure safe separation. Agent-based models for conflict resolution in dense UAS operations were introduced in \cite{doi:10.2514/1.D0119}. Geofencing strategies for managing low-altitude UAS traffic in urban environments were proposed in \cite{9298470, app12020576}. A vector-field-based method for deconfliction of AAM vehicles was presented in \cite{10311214}. Rapidly-exploring random trees (RRT) was utilized in \cite{9594489} for path planning in urban settings with static and dynamic obstacles. Stochastic optimal control approaches were formulated in \cite{MATSUNO201577, 6579753} to address conflict resolution under uncertain wind conditions. A method for identifying high-collision-risk airspace hot spots was proposed in \cite{7909027}, while mid-air collision risks in AAM operations were investigated in \cite{10598220}.

Several methods have been proposed for in-flight emergency risk mitigation \cite{drones9020141}. Neural networks have performed landing site identification in contingency planning frameworks \cite{10488310, castagno2021map}. Dubins path solvers were used in \cite{tekaslan2025_v2v} to generate emergency road landing trajectories. Dubins paths, combined with RRT, were also utilized in \cite{10858596} for trajectory generation. As an alternative to geometric solvers, search-based algorithms have been applied to emergency landing planning \cite{castagno2021map, tekaslan_search, 9739975}, offering greater flexibility in handling complex constraints. In addition, prognostic-based emergency risk mitigation was demonstrated using Markov Decision Processes (MDP) \cite{akinola2025markov, sharma2024risk} to support decision-making under uncertainty.

Despite advancements in emergency trajectory planning and airspace deconfliction, integration of the two remains limited. Most emergency landing studies focus on static obstacles or ground risk, largely ignoring airspace complexity and its operational consequences. In contrast, air deconfliction methods typically assume nominal aircraft performance, full control authority, and compliance with flight plans. The challenge of assuring separation during emergencies when an aircraft has degraded performance has not been adequately addressed. This study aims to fill this gap by jointly addressing real-time contingency landing planning and reduced airspace disruption, with particular emphasis on congested urban environments.

Within this scope, risk is quantified by the time an emergency landing path spends within or near active air traffic corridors and/or no-fly zones. Reducing this exposure lowers conflict likelihood, and leads to safer landings and overall safer airspace. Disturbance to nominal airspace operations is reduced with an airspace-aware emergency landing planner that explicitly steers trajectories around traffic corridors and no-fly zones, extending the discrete search-based contingency planner avoiding Q densely populated areas presented in \cite{tekaslan_search}. Air traffic occupancy is derived from Automatic Dependent Surveillance–Broadcast (ADS-B) data, which primarily capture centrally managed Instrument Flight Rules (IFR) traffic. In contrast, Visual Flight Rules (VFR) traffic is less reliably observed because low-altitude operations may fall outside ADS-B coverage, and some operators, including military aircraft, are exempt from broadcasting. To address these limitations, aeronautical charts are incorporated to model low-altitude corridors, and hierarchical computational geometry is employed to augment proximity-based risk metrics within these regions. The trajectory planner concurrently minimizes exposure to congested airspace and high ground population density. Its performance is assessed with engine-out Cessna 182 (C182) scenarios in Washington, D.C. airspace, incorporating real-world ADS-B traffic and steady wind conditions. Feasibility is assessed through dynamic simulations, and all planner implementations and datasets are available on GitHub. \footnote{\raggedright \url{https://github.com/tekaslan/GGS-ACLM}}. 

This paper contributes a novel methodology for processing air traffic and airspace use data into collision risk heatmaps and exclusion zones. Airspace and ground risk are jointly minimized in emergency landing planning.
A novel airspace deconfliction impact metric is defined for emergency landing site selection.  A statistical analysis and graphical presentation of complexity and constraint satisfaction for emergency landing planning in congested airspace is presented. 

This paper is organized as follows. Section \ref{sec:prelim} summarizes fixed-wing aircraft gliding performance and emergency landing planning with search and Dubins solvers.  Section \ref{sec:method} presents airspace traffic density determination using ADS-B data and a computational geometry framework to generate proximity-based heatmaps of urban air corridors and prohibited areas. Landing site selection with air traffic disruption metrics and emergency landing trajectory generation for air deconfliction are detailed in Section \ref{sec:method}. Section~\ref{sec:results} benchmarks risk and computation emergency landing planning metrics for the C182 under airspace-only and joint airspace–ground risk scenarios. Sections \ref{sec:discussion} and \ref{sec:conclusion} conclude the paper.
\section{Preliminaries}
\label{sec:prelim}
This section summarizes fundamentals of search-based path planning and the gliding aircraft flight envelope as background. 

\subsection{Gradient-Guided Search and Ground Risk Minimization}
For path planning, tree-search incrementally explores the future states of an agent based on a cost function by branching over reachable options to identify a a path that satisfies maneuver constraints. This paper builds upon the gradient-guided search for emergency landing planning presented in \cite{tekaslan_search, tekaslan_search_icra}.
Define aircraft state as  $s = (\varphi, \lambda, h, \chi)$ where $\varphi \in \Phi$, $\lambda \in \Lambda$, $h \in \mathbb{R}_{\geq 0}$, and $\chi \in [0, 2\pi)$ respectively correspond to latitude, longitude, altitude above mean sea level (MSL), and course angle. Valid sets of latitude and longitude are denoted by $\Phi$ and $\Lambda$. Search space $\mathcal{S}$ is defined per \eqref{eq:search_space}
\begin{equation}
\begin{aligned}
    \mathcal{S} = \left\{ (\varphi, \lambda, h, \chi) \mid \varphi \in \Phi, \lambda \in \Lambda, \ h \in [0, \infty),\chi \in [0, 2\pi) \right\}
    \end{aligned}
    \label{eq:search_space}
\end{equation}
Let the initial emergency state and target destination (approach fix) states be $s_0 \in \mathcal{S}$ and $s_N \in \mathcal{S}$, respectively. A contingency landing path $\mathcal{P} \subset \mathcal{S}$ consists of a sequence of states from $s_0$ to $s_N$ where $\mathcal{P} = \{s_0, s_1, \cdots, s_N\}$.

Given the gliding aircraft flight envelope $\mathbb{E}$, action set $\mathcal{A}$, initial and goal states $s_0$ and $s_N$, cost function $f(s)$, and steady wind parameters $(v_w, \chi_w)$, the four-dimensional (4D) search-based path planning problem is formulated as
\begin{equation}
    \mathcal{P}_s(s_0, s_N) = \left\{ s \mid s = \mathcal{T}(\mathbb{E}, \mathcal{A}, f(s), s_0, s_N, v_w, \chi_w) \right\},
    \label{eq:tree_search}
\end{equation}
where $\mathcal{P}_s$ is a non-ascending emergency landing trajectory generated by the tree search planner. 

Function $f(s)$ estimates the cost of each child state and action and does not include cumulative cost of preceding states leading to $s$. In that sense, it is a greedy cost that was proven effective for real-time performance and ground risk avoidance in \cite{tekaslan_search}. $f(s)$ was originally constructed from four terms: (1) a gradient prioritizing states along the optimal descent path angle, $h_{d,1}:\mathcal{S} \rightarrow [0,1]$; (2) a gradient guiding search toward the goal via minimum remaining traversal, $h_{d,2}:\mathcal{S} \rightarrow [0,1]$; (3) a gradient aligning heading with desired course near the goal, $h_{\chi}:\mathcal{S} \rightarrow [0,1]$; and (4) a ground risk term discouraging flight over densely populated areas, $h_{p}:\mathcal{S} \rightarrow [0,1]$.

\begin{figure}[t!]
    \centering
    \includegraphics[width=\linewidth]{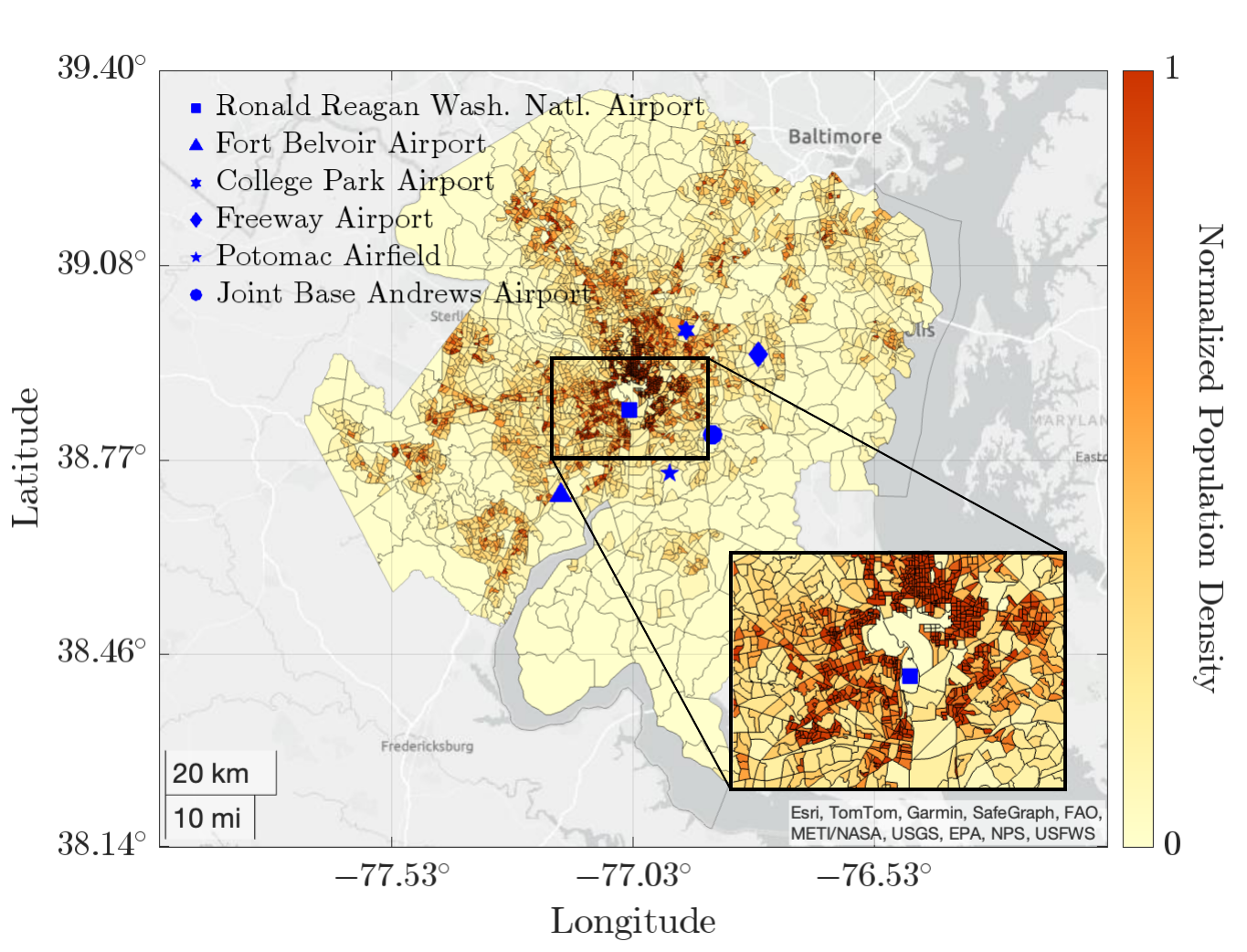}
    \caption{Normalized population density around Washington, D.C.}
    \label{fig:population_density}
\end{figure}

Figure \ref{fig:population_density} shows normalized population density \cite{Census2020DECENNIALPL2020.P1} as a function of geodetic coordinates near Washington, D.C. with an area-weighted mean normalized density of $\mu_{\eta} = 0.1374$.
The normalized overflown population density as a function of flight time is denoted by $\eta : [0, T] \rightarrow [0, 1]$. To ensure balanced state expansion in total cost,  ground risk component $h_p(s)$ is normalized by the maximum value of the integrand $\eta_{w,\max} < 1$. Arrival time of a state $s$ along the trajectory $\mathcal{P}$ is given by the mapping $T: \mathcal{S} \rightarrow \mathbb{R}_{\geq 0}$. The time-averaged, normalized overflown population cost between two consecutive states $s_i$ and $s_j$ is defined per \eqref{eq:ground_risk}. Adaptive weight functions $w_{p,1}(t)$ and $w_{p,2}(t)$ are used to downscale ground risk cost proportional to altitude and remaining traversal.
\begin{equation}
    h_p(s) = \frac{1}{\eta_{w,\max} [T(s_j) - T(s_i)]} \int_{T(s_i)}^{T(s_j)} w_{p,1}(t)w_{p,2}(t)\eta(t) \, dt
    \label{eq:ground_risk}
\end{equation}

The remaining cost terms are not presented in this paper for brevity. Explicit formulations and boundedness proofs of each cost term are in \cite{tekaslan_search}. This paper formulates total cost $f(s)$ in the form $f(s) = g(s) + h(s)$ where $g(s)$ is the cumulative path cost quantifying airspace exposure, and $h(s)$ is a heuristic cost including greedy ground risk. Cost functions to minimize the need for airspace deconfliction are defined in Section \ref{sec:deconfliction_search}.

\subsection{Gliding Fixed-wing Aircraft Performance Under Wind}
Glide performance of wing-lift aircraft including best and steepest glide is well established \cite{tekaslan2025_v2v, tekaslan_search, Raymer2012}. To ensure safe and dynamically feasible emergency landing paths, airspeed $v_a \in \mathbb{R}^+$ must remain within operational bounds. Best-glide speed $v_{bg}$ maximizes aerodynamic lift-to-drag ratio thus range but is prone to stall under propulsion loss or variable winds. $v_{bg}$ is therefore adopted as the lower bound of the airspeed envelope. As flaps are essential for reducing stall and ground speeds, the upper bound is defined by maximum flap-extended airspeed $v_{FE}$ above which structural loads may cause damage. The reference airspeed $v_a^\star$ is chosen as the midpoint between $v_{bg}$ and $v_{FE}$ to maximize controllability safety margins:
\begin{equation}
    v_a^\star = \frac{v_{bg} + v_{FE}}{2}
    \label{eq:vref}
\end{equation}

Airspeed is defined in the wind frame \cite{beard_mclain}, while landing control acts in the inertial frame. In engine-out flight, modifying pitch attitude is the primary means of regulating airspeed because it governs the aircraft’s energy exchange. To support airspeed boundedness with formal guarantee, a viability-based parametric optimization across discrete maneuvers $\sigma = (\Delta\chi, \mathbf{v}_w; \mathbf{v}_a) \in \Sigma$ was conducted for an engine-out Cessna 182 \cite{tekaslan_viability} where $\Sigma$ is the admissible maneuver set. Here, $\Delta\chi \in [-2\pi,2\pi]$ is a course change (i.e., lateral maneuver), and $\mathbf{v}_w, \mathbf{v}_a \in \mathbb{R}^3$, are, respectively, wind and airspeed vectors. The viability-constrained optimal control problem minimizes acceleration $\dot{v}_a$ over the maneuver horizon $T(\sigma)$:
\begin{equation}
\begin{gathered}
\gamma_g^\star(\sigma) =
\operatorname*{argmin}
\int_{0}^{T(\sigma)} \dot{v}_a^2\,dt \\
\text{s.t. } \gamma_g \in \Gamma_{\mathcal{V}}, \quad \forall t \in [0,T(\sigma)],
\end{gathered}
\label{eq:vc_opt_compact}
\end{equation}
where $\Gamma_{\mathcal{V}}$ is the viable ground-reference flight path angle set. The resulting flight path angle solutions (i.e., longitudinal maneuver) populate a look-up table for real-time trajectory planning with formal airspeed boundedness.
\section{Problem Statement}
This paper incorporates proactive mid-air collision avoidance into aircraft contingency landing planning by minimizing air traffic corridor exposure. This problem is challenging due to two primary factors: the complexity of the airspace operational environment and the need for real-time plan computation. First, emergency landing planning must account for off-nominal aircraft performance under varying wind conditions to ensure dynamic feasibility. Also, associated risks to other aircraft in the airspace as well as risk to people on the ground must be modeled and integrated in the planner. Airspace risk modeling requires accurate evaluation of risk metrics associated with different airspace structures including restricted zones, aircraft operating under instrument flight rules with known flight plans, and other aircraft flying under visual flight rules. Navigating through congested airspace with reduced maneuverability poses a highly constrained problem in route planning. The second major challenge is computational. A contingency landing planner may need to be triggered at any point during flight, necessitating the generation of a feasible and safe landing trajectory within strict runtime constraints. This requirement is further complicated by a potentially large search space with constraints. As a result, the planner must balance runtime efficiency and search-space exploration to support practical deployment. For this study, two assumptions are made: reachability to at least one landing site is confirmed, and the steady wind field is known.
\section{Methodology}
\label{sec:method}
The airspace near an urban airport is shared by arrival and departure traffic, rotorcraft, and  transiting aircraft. All aircraft in Class B airspace must follow prescribed routes while avoiding no-fly zones. This section presents a heatmap risk modeling framework for mapped airspace structures  illustrated with examples from the Washington, D.C, region.

\subsection{Airport Departure/Arrival Traffic Corridors}
\label{sec:a/d_traffic_corridors}
\begin{figure}[t!]
    \centering
    \includegraphics[width=\linewidth]{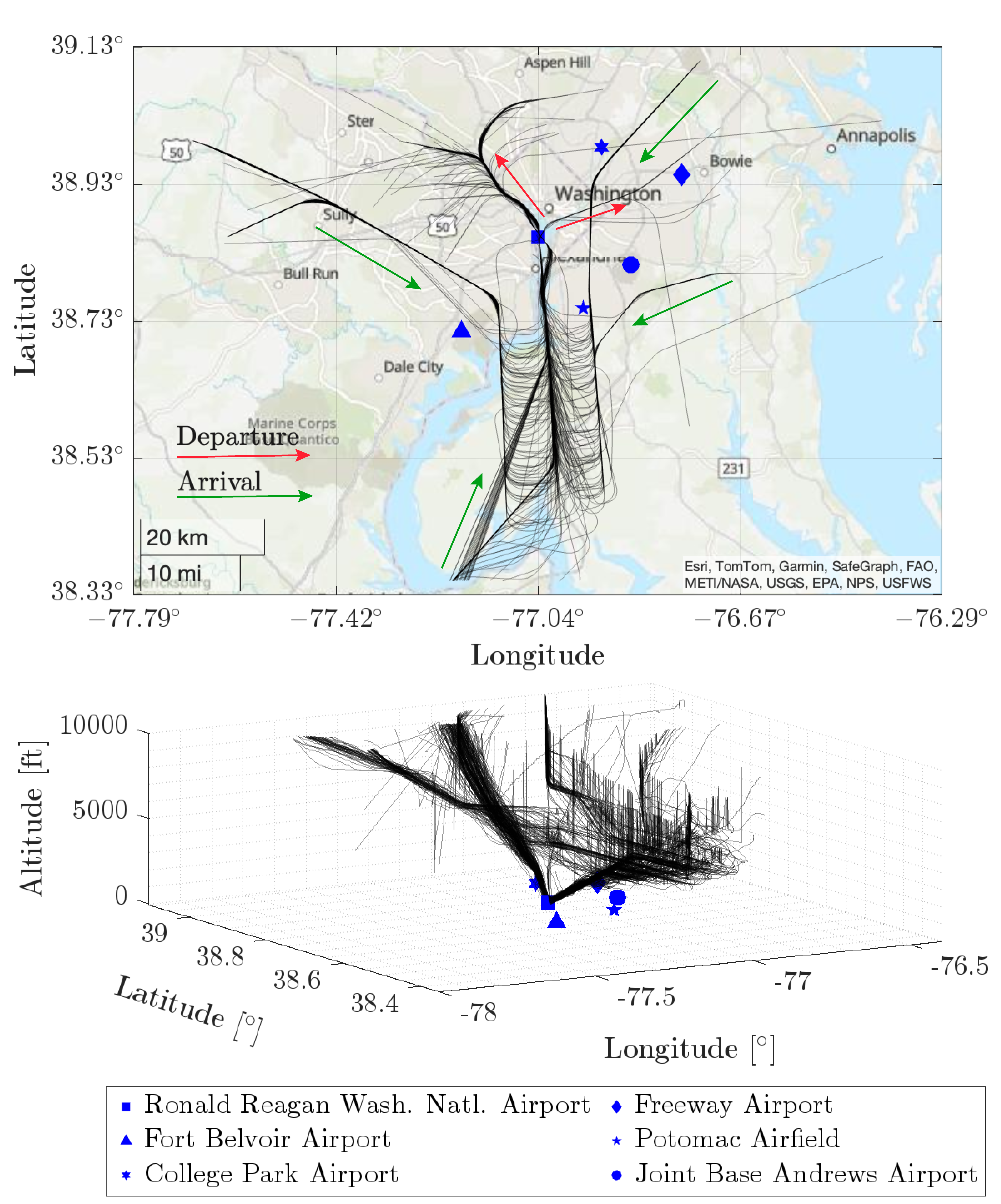}
    \caption{Ronald Reagan Washington National Airport, North flow ADS-B flight trajectories - Jan. 29, 2025, from 12:07 p.m. to 8:49 p.m. local time.}
    \label{fig:dca_traffic}
\end{figure}

Aircraft today must be equipped with Automatic Dependent Surveillance - Broadcast (ADS-B) with limited exceptions. The OpenSky Network \cite{opensky} offers historical ADS-B data that enables statistical characterization of aircraft occupancy in a given airspace volume. Figure \ref{fig:dca_traffic} shows the flight trajectories below 10000 ft MSL to and from Ronald Reagan Washington National Airport (DCA) on January 29, 2025, a notorious date due to an aircraft-rotorcraft collision. The dataset includes all flights from the onset of North flow operations at 12:07 p.m. until 8:49 p.m., the local time at which the mid-air collision occurred. The dataset extends approximately 60 NM East-West and 45 NM North-South. During this period, air traffic arriving at DCA predominantly follows the published terminal routes at varying altitudes, while the dispersion of trajectories near the final turn indicates vectoring by ATC. It can be inferred that both the terminal arrival routes and the initial departure climb paths are highly likely to contain air traffic thus are represented with high cost in the airspace risk heatmap.

\begin{figure}[hbt!]
    \centering
    \includegraphics[width=0.9\linewidth]{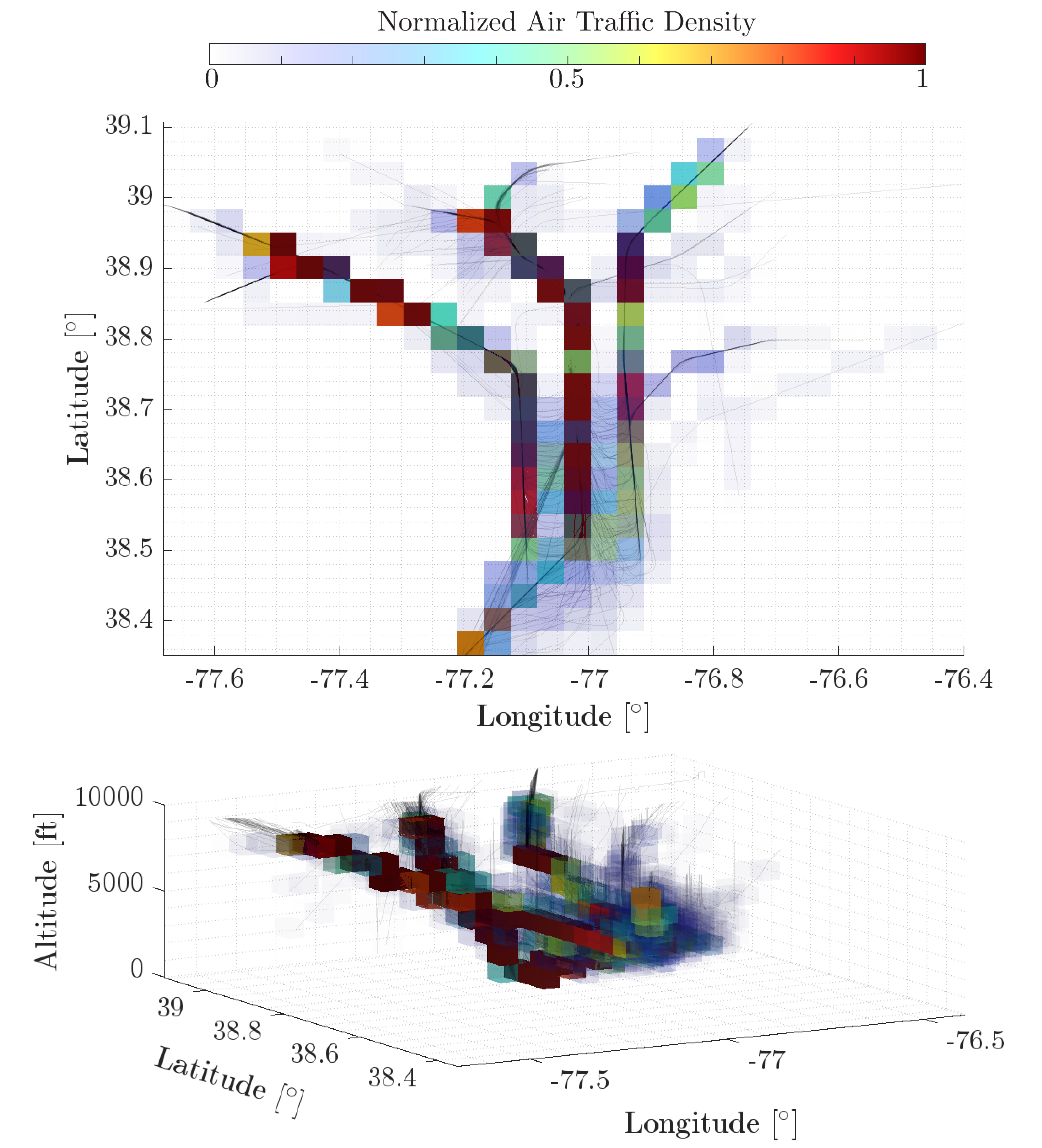}
    \caption{Ronald Reagan Washington National Airport air traffic heatmap.}
    \label{fig:dca_traffic_cost}
\end{figure}

In this study, the airspace is uniformly discretized into a structured grid of three-dimensional cells where $N_{\varphi} = 30$, $N_{\lambda} = 30$, and $N_{h} = 10$. Regions with high departure and arrival traffic density are identified using a static ADS-B dataset. Specifically, trajectory segments traversing each cell are used to estimate air traffic density within that cell. Let airspace volume $\mathbb{A} \subset \mathbb{R}^3$ be uniformly discretized into cells, where the number of divisions along  latitude, longitude, and altitude dimensions are denoted $N_\varphi$, $N_\lambda$, and $N_h$, respectively:
\begin{equation}
\mathbb{A} = \bigcup_{i=1}^{N_{\varphi}} \bigcup_{j=1}^{N_{\lambda}} \bigcup_{k=1}^{N_h} C_{i,j,k}
\label{eq:cell}
\end{equation}
Each cell $C_{i,j,k}$ spans a fixed spatial extent per \eqref{eq:cell2}.
\begin{equation}
C_{i,j,k} = [\varphi_i, \varphi_{i+1}) \times [\lambda_j, \lambda_{j+1}) \times [h_k, h_{k+1})
\label{eq:cell2}
\end{equation}
As defined in \eqref{eq:adsb_traj}, let $\mathbb{T}$ denote the set $m \in \mathbb{Z}^+$ of recorded flight trajectories $\tau$ obtained from ADS-B data. Each trajectory $\tau \in \mathbb{T}$ is a sequence of three-dimensional positions $\mathbf{p} \in \mathbb{R}^3$, where each position is given by a projection $\mathbf{p} = \pi(s)$ of a full aircraft state $s \in \mathcal{S}$. The projection function $\pi: \mathcal{S} \rightarrow \mathbb{R}^3$ extracts the spatial components of $s$ such that $\mathbf{p} = [\varphi\;\lambda\;h]^\intercal$. The number of discrete positions on a trajectory $n\in\mathbb{Z}^+$ is identified based on total trajectory length $\lVert \tau \rVert \in \mathbb{R}^+$ and step length $d_{\text{step}} = 100$ ft such that $d_{\text{step}} \ll \lVert \tau \rVert$.
\begin{equation}
\begin{aligned}
\mathbb{T} = \{\tau_m \mid \tau_m = \{ \mathbf{p}_{m,n} \}, n = \ceil{\lVert \tau \rVert/ d_{\text{step}}}, d_{\text{step}} \in \mathbb{R}^+\}
\end{aligned}
\label{eq:adsb_traj}
\end{equation}
Let $\mathbb{K}$ be the set of air traffic density $\kappa_{i,j,k}$ of $C_{i,j,k}$:
\begin{equation}
\mathbb{K} = \{\kappa_{i,j,k} \mid \kappa \in \mathbb{Z}_{\geq 0}\}
\label{eq:air_traffic_density_set}
\end{equation}
Air traffic density $\kappa$ is the sum of the number of discrete positions over all trajectories that lie inside a given cell:
\begin{equation}
\kappa_{i,j,k} = \sum_{a=1}^{m} \sum_{b = 1}^{n} \delta(\mathbf{p}_{a,b} \mid C_{i,j,k})
\label{eq:air_traffic_density},
\end{equation}
where indicator function $\delta:\mathbb{R}^3\times \mathbb{A} \rightarrow \{0,1\}$ is defined as
\begin{equation}
\delta(\mathbf{p} \mid C) = 
\begin{cases}
1, & \text{if } \mathbf{p} \in C\\
0, & \text{otherwise}
\end{cases}
\;.
\label{eq:indicator}
\end{equation}
 $\kappa$ can be considered a scaled probability of aircraft occupancy in a given cell. 
 The airport traffic heatmap is specified by $\overline{\mathbb{K}}$ and is a set of normalized air traffic density of discrete cells within the considered airspace per \eqref{eq:heatmap_traffic}.
\begin{equation}
    \overline{\mathbb{K}} = \{\bar{\kappa}_{i,j,k} \mid \bar{\kappa} = \kappa/\kappa_{\max}, \kappa_{\max} = \sup \mathbb{K}\}
\label{eq:heatmap_traffic}
\end{equation}
Air traffic density as a function of position $\kappa:\mathbb{R}^3 \rightarrow [0,1]$ is further defined as follows:
\begin{equation}
    \kappa(\mathbf{p}) = \sum_{i=1}^{N_{\varphi}} \sum_{j=1}^{N_{\lambda}} \sum_{k=1}^{N_{h}} \bar{\kappa}_{i,j,k} \delta(\mathbf{p} \mid C_{i,j,k}).
\end{equation}
This formulation yields exactly one nonzero term in the summation due to airspace partitioning. Figure \ref{fig:dca_traffic_cost} shows the resulting heatmap for traffic near DCA airport. As previously discussed, arrival and departure paths correspond to high-cost volumes that should be avoided.
Grid cells with positive air traffic density $\kappa_{i,j,k} > 0$ are sparse in $\mathbb{A}$ as seen in Figure \ref{fig:dca_traffic_cost}, occupying only $9.3\%$ of the considered airspace. Mean normalized air traffic density $\mu_{\kappa}$ is 0.014 per \eqref{eq:mean_kappa}.
\begin{equation}
    \mu_{\kappa} = \frac{1}{N_{\varphi} + N_{\lambda} + N_{h}}\sum_{i=1}^{N_{\varphi}} \sum_{j=1}^{N_{\lambda}} \sum_{k=1}^{N_{h}} \bar{\kappa}_{i,j,k} = 0.014
    \label{eq:mean_kappa}
\end{equation}

\subsection{Polyhedral Airspace Modeling}
\begin{figure}[t!]
    \centering
        \includegraphics[width=0.9\linewidth]{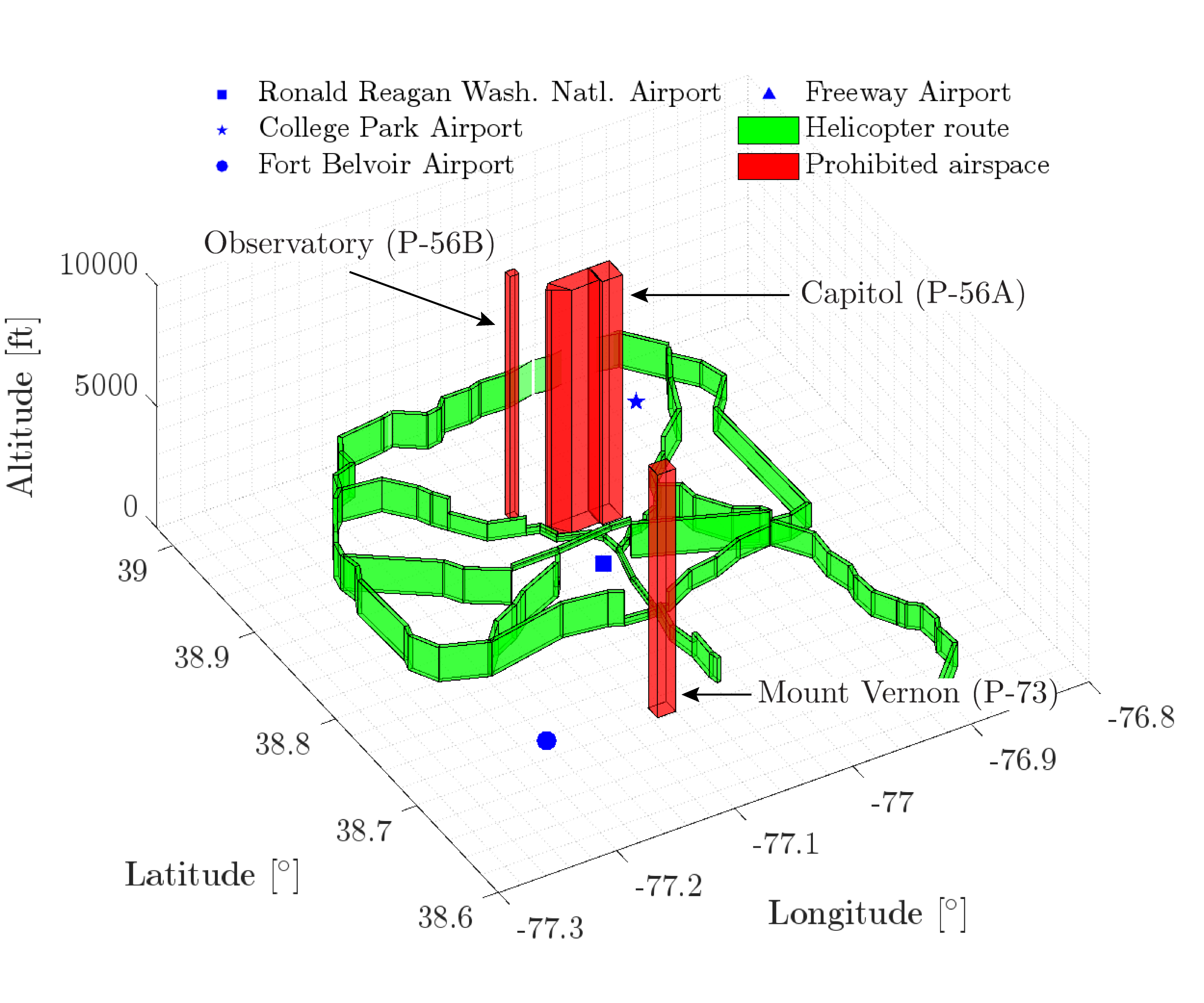}
        \caption{DCA area helicopter routes and no-fly zones modeled as polyhedra.}
        \label{fig:dc_polyhedron}
\end{figure}

At the time of this study, urban areas include designated air traffic corridors for helicopters and general aviation aircraft, particularly those operating under visual flight rules. 
Airspace volumes can also be designated with a temporary flight restriction or permanently prohibited for air traffic. For path planning, these regions can be considered obstacles. Each corridor is designated with an altitude range. Both air corridors and prohibited volumes are represented with three-dimensional (3D) convex polyhedra in this work based on aeronautical charts \cite{skyvector}. Figure \ref{fig:dc_polyhedron} depicts helicopter corridors and prohibited zones in the Washington D.C. area as polyhedra.

Let polyhedron sets that contain $n_u$ urban air corridors and $n_n$ no-fly zones respectively be
$\mathbb{P}_u$ and $\mathbb{P}_n$ per \eqref{eq:polyhedra_sets}.
\begin{equation}
\begin{aligned}
    \mathbb{P}_u &= \{P_1, \cdots, P_{n_u} \mid {n_u} \in \mathbb{Z}_{\geq 0}\}\\
    \mathbb{P}_n &= \{P_1, \cdots, P_{n_n} \mid {n_n} \in \mathbb{Z}_{\geq 0}\}
    \end{aligned}
    \label{eq:polyhedra_sets}
\end{equation}
Here, $P$ is a 3D polyhedron. An arbitrary position sample $\mathbf{p}$ on an emergency landing trajectory $\mathcal{P}_s$ may lie inside a corridor, on one of its faces, or outside all corridors, in which case it is closest to the corridor with the minimum distance per Eq. \ref{eq:dmin}. Therefore, mid-air conflict risk at position $\mathbf{p}$ can be quantified based on $d_{\min}: \mathbb{R}^3\times \mathbb{P} \rightarrow \mathbb{R}_{\geq0}$.
\begin{equation}
d(\mathbf{p} \mid P) = 
    \begin{cases}
        0, & \mathbf{p} \in P\\
        d_{\min}(\mathbf{p} \mid P), & \mathbf{p} \notin P
    \end{cases}
    \label{eq:dmin}
\end{equation}
A 3D point-in-polyhedron algorithm is defined to determine if point $\mathbf{p} \in P$. A hierarchical computational geometry method is employed to compute minimum distance $d_{\min}(\mathbf{p} \mid P): \mathbb{R}^3 \times \mathbb{P} \rightarrow \mathbb{R}^+$ for cases where $\mathbf{p} \notin P$. A formal definition of a polyhedron is first provided, and both methods are then described mathematically. Note that position vectors are transformed from the Latitude-Longitude-Altitude frame to an Earth-Centered Earth-Fixed frame for all vector operations.

\begin{figure}[hbt!]
    \centering
    \includegraphics[width=.75\linewidth]{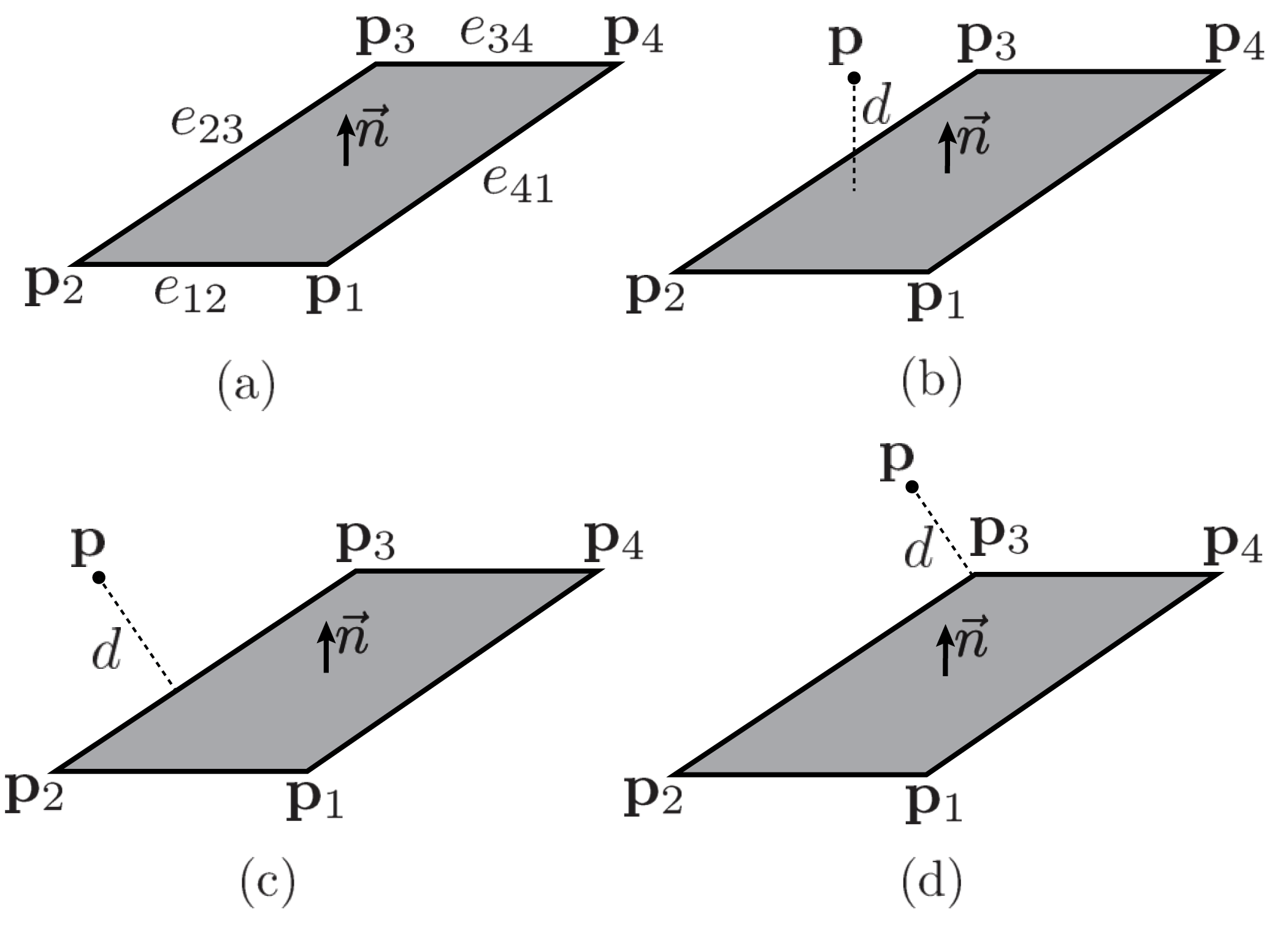}
    \caption{A polyhedron face defined by its edges, vertices, and normal vector, along with the relative positions of an arbitrary point with respect to the face.}
    \label{fig:face}
\end{figure}

Let a polyhedron enclosed by six faces be denoted $P = \{f_i \mid i = 1, \ldots, 6\}$. Each face $f$, with an associated normal vector $\vec{n}$, is a planar region defined by four edges as illustrated in Figure \ref{fig:face}(a). An edge $e_{ij} = e(\mathbf{p}_i, \mathbf{p}_j)$ is a straight line segment connecting points $\mathbf{p}_i$ and $\mathbf{p}_j$. 
The unit face normal can be found with a cross product of edge vectors as in \eqref{eq:face_normal}.
\begin{equation}
    \vec{n} = \frac{(\mathbf{p}_2 - \mathbf{p}_1)\times (\mathbf{p}_4 - \mathbf{p}_1)}{\lvert \mathbf{p}_2 - \mathbf{p}_1 \rvert\lvert \mathbf{p}_4 - \mathbf{p}_1 \rvert}
    \label{eq:face_normal}
\end{equation}

\subsection{Point-in-Polyhedron Detection}
A geometric method is implemented to determine whether an arbitrary point $\mathbf{p}$ lies within a given polyhedron. Let $\mathbf{c}$ denote the centroid of face $f$, computed per \ref{eq:centroid}.
\begin{equation}
    \mathbf{c} = \frac{1}{4} \sum_{i=1}^4 \mathbf{p}_i
    \label{eq:centroid}
\end{equation}
Let $\vec{u}_i = \mathbf{p} - \mathbf{c}_i$ be a vector, pointing from the centroid $\mathbf{c}_i$ of face $f_i$ to the query point $\mathbf{p}$. To determine whether $\mathbf{p} \in P$, the dot products $\vec{n}_i \cdot \vec{u}_i$ for each face $f_i$ is evaluated, where $\vec{n}_i$ is the outward-pointing unit normal. If all projections are non-positive, i.e., $ \vec{n}_i \cdot \vec{u}_i \leq 0$, $\forall i$, the point lies inside or on the boundary of the polyhedron. Conversely, if any projection is positive, the point is deemed to be situated outside. Let $o:\mathbb{R}^3\times\mathbb{P} \rightarrow \{0, 1\}$ be another indicator function to identify whether given polyhedron encompasses $\mathbf{p}$ per \eqref{eq:pip}.
\begin{equation}
    o(\mathbf{p} \mid P) = 
    \begin{cases}
        1, & \vec{n}_i \cdot \vec{u}_i \leq 0,\ \forall i \\
        0, & \exists i \text{ such that } \vec{n}_i \cdot \vec{u}_i > 0
    \end{cases}
    \label{eq:pip}
\end{equation}
If $\mathbf{p}\in P$ such that $o(\mathbf{p} \mid P) = 1$, $\mathbf{p}$ is penalized with a positive finite cost for path planning purposes as it dwells inside an air traffic corridor. Otherwise, if $o(\mathbf{p} \mid P) = 0$, the airspace cost of $\mathbf{p}$ is determined based on the minimum distance value $d_{\min}(\mathbf{p} \mid P)$.
\begin{figure}[hbt!]
    \centering
    \includegraphics[width=0.85\linewidth]{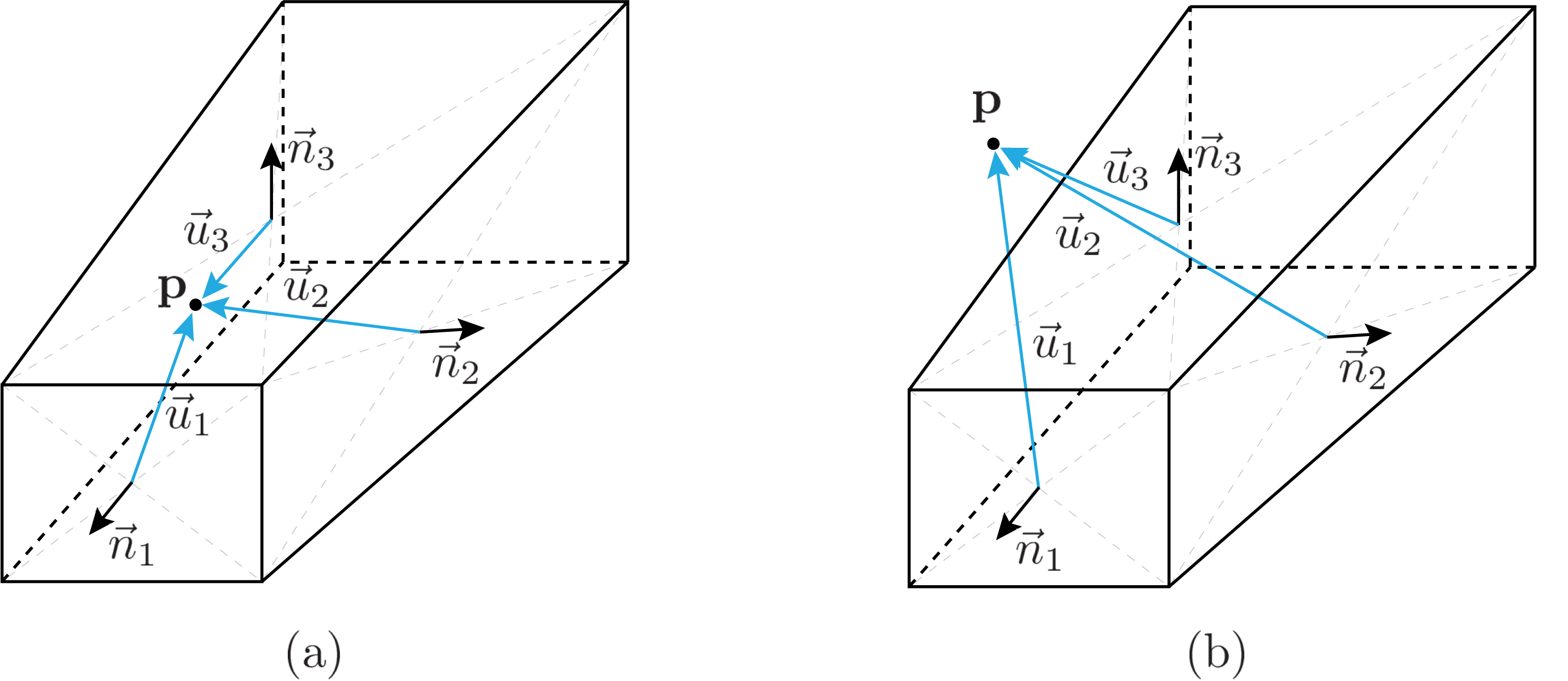}
    \caption{The relative positions of a point with respect to a polyhedron.}
    \label{fig:pip}
\end{figure}
Figure~\ref{fig:pip} illustrates the evaluation of \eqref{eq:pip} using a simplified polyhedron. In the left subfigure (a), the query point $\mathbf{p}$ lies inside the polyhedron, as each vector $\vec{u}_i$ has a component in the direction opposite to the corresponding outward normal $\vec{n}_i$, i.e., $\vec{n}_i \cdot \vec{u}_i \leq 0$, $\forall i$. In contrast, the right subfigure (b) shows $\mathbf{p}$ located outside the volume, which is indicated by a positive dot product $\vec{n}_3 \cdot \vec{u}_3 > 0$, meaning that $\vec{u}_3$ has a component pointing the same direction as $\vec{n}_3$.

\subsection{Point-to-Polyhedron Minimum Distance}
\begin{figure}[t!]
    \centering
    \includegraphics[width=.7\linewidth]{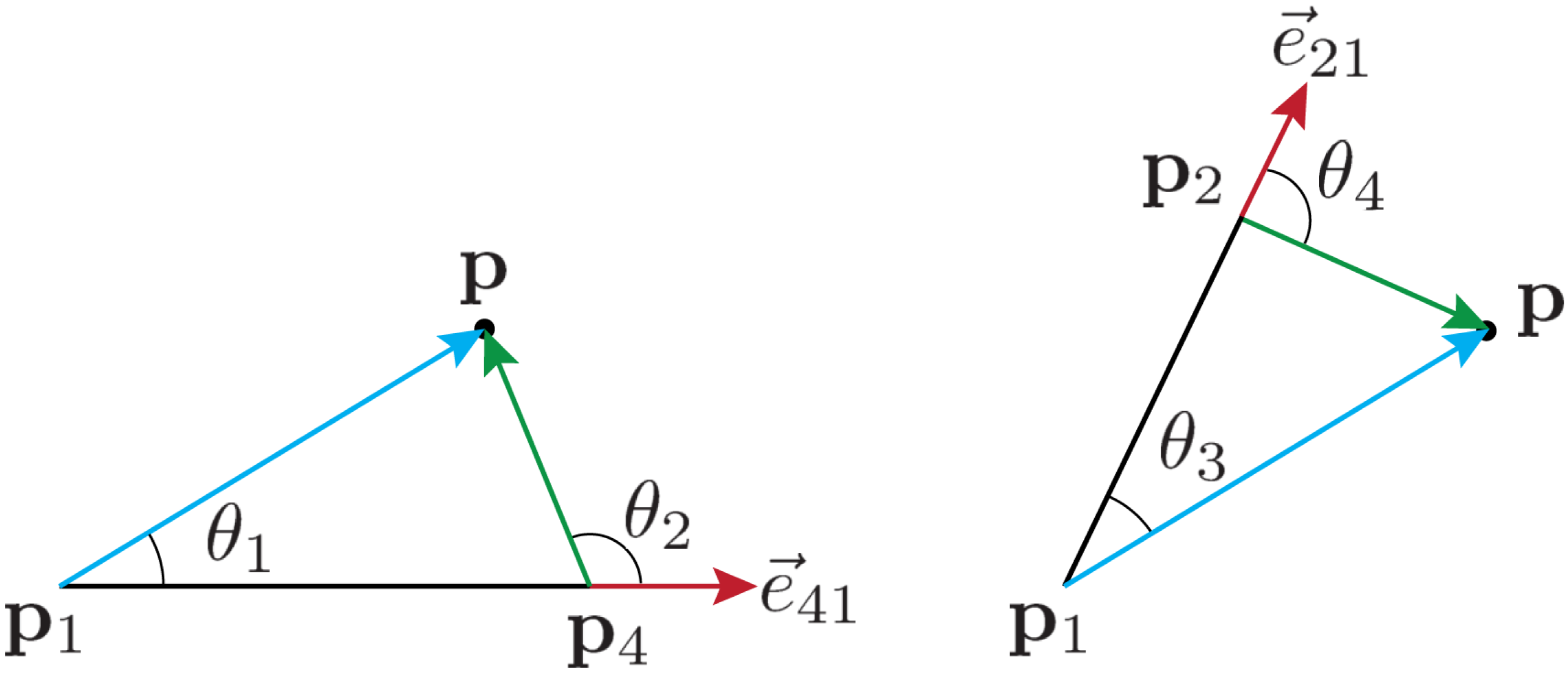}
    \caption{Angles between surface edges and point-to-vertex vectors.}
    \label{fig:enter-angle}
\end{figure}

As discussed above, if $\mathbf{p} \notin P$ for all $P \in \mathbb{P}$, the minimum distance from $\mathbf{p}$ to the set of polyhedra must be computed to evaluate airspace risk. A hierarchical geometric method is employed to determine the relative position of $\mathbf{p}$ with respect to each polyhedron in $\mathbb{P}$. Depending on the relative positions of $\mathbf{p}$ and $P$, the minimum distance may correspond to the shortest distance from the point to a face, an edge, or a vertex, as respectively shown in Figure~\ref{fig:face}~(b), (c), and (d). Now, consider a plane containing face $f$, expressed as:
\begin{equation}
    ax + by + cz - k = 0,
    \label{eq:plane}
\end{equation}
where $\vec{n} = [a\; b\; c]^\intercal$ and $[a\; b\; c\; k]^\intercal \in \mathbb{R}^4$. If query point $\mathbf{p}$ is contained within the region bounded by the edges of face $f$ along the direction of $\vec{n}$, then its orthogonal projection onto the plane intersects the face. This condition can be evaluated using angular relationships, as shown in Figure \ref{fig:enter-angle} and formulated in \eqref{eq:angle_conditions} where $\varepsilon > 0$ is a small number for numerical stability.
\begin{equation}
\begin{aligned}
    \theta_1 &= \cos^{-1}\left(\frac{( \mathbf{p} - \mathbf{p}_1) \cdot \vec{e}_{41}}{\lvert \mathbf{p} - \mathbf{p}_1\rvert \lvert \vec{e}_{41} \rvert + \varepsilon} \right) \leq \frac{\pi}{2}\\
    \theta_2 &= \cos^{-1}\left(\frac{( \mathbf{p} - \mathbf{p}_4) \cdot \vec{e}_{41}}{\lvert \mathbf{p} - \mathbf{p}_4\rvert \lvert \vec{e}_{41} \rvert+ \varepsilon} \right) \geq \frac{\pi}{2} \\
    \theta_3 &= \cos^{-1}\left(\frac{( \mathbf{p} - \mathbf{p}_1) \cdot \vec{e}_{21}}{\lvert \mathbf{p} - \mathbf{p}_1\rvert \lvert \vec{e}_{21} \rvert + \varepsilon} \right) \leq \frac{\pi}{2}\\
    \theta_4 &= \cos^{-1}\left(\frac{( \mathbf{p} - \mathbf{p}_2) \cdot \vec{e}_{21}}{\lvert \mathbf{p} - \mathbf{p}_2\rvert \lvert \vec{e}_{21} \rvert+ \varepsilon} \right) \geq \frac{\pi}{2}
\end{aligned}
\label{eq:angle_conditions}
\end{equation}
If the conditions in \eqref{eq:angle_conditions} hold, the point-to-face minimum distance (see Figure \ref{fig:face} (b)) is given by the orthogonal distance from $\mathbf{p}$ to the plane per \eqref{eq:point_to_plane}.
\begin{equation}
    d_{\min}(\mathbf{p}\mid P) = \frac{\lvert \vec{n} \cdot \mathbf{p} + k \rvert}{\lvert \vec{n} \rvert} = \lvert \vec{n} \cdot \mathbf{p} + k \rvert
    \label{eq:point_to_plane}
\end{equation}
If $\mathbf{p}$ lies outside the region on the plane bounded by the face edges, the minimum distance is determined based on its position relative to the face's edges and vertices. The following cases are considered:
\begin{enumerate}
    \item $\mathbf{p}$ lies between vertices $\mathbf{p}_1$ and $\mathbf{p}_4$ but not between $\mathbf{p}_1$ and $\mathbf{p}_2$. Then,
    \begin{equation}
        d_{\min}(\mathbf{p}\mid P) = \frac{\lvert \vec{e}_{41} \times (\mathbf{p} - \mathbf{p}_1) \rvert}{\lvert \vec{e}_{41}\rvert}.
        \label{eq:point_to_edge1}
    \end{equation}

    \item $\mathbf{p}$ lies between vertices $\mathbf{p}_1$ and $\mathbf{p}_2$ but not between $\mathbf{p}_1$ and $\mathbf{p}_4$. Then,
    \begin{equation}
        d_{\min}(\mathbf{p}\mid P) = \frac{\lvert \vec{e}_{21} \times (\mathbf{p} - \mathbf{p}_1) \rvert}{\lvert \vec{e}_{21} \rvert}.
        \label{eq:point_to_edge2}
    \end{equation}

    \item $\mathbf{p}$ lies outside both edges defined by $e_{41}$ and $e_{12}$. Then,
    \begin{equation}
        d_{\min}(\mathbf{p}\mid P) = \inf \left\{ \lVert \mathbf{p} - \mathbf{p}_i\rVert \rvert \;\middle|\; i = 1, \dots, 4 \right\}.
        \label{eq:point_to_vertex}
    \end{equation}
\end{enumerate}

\begin{figure}[hbt!]
    \centering
    \includegraphics[width=.8\linewidth]{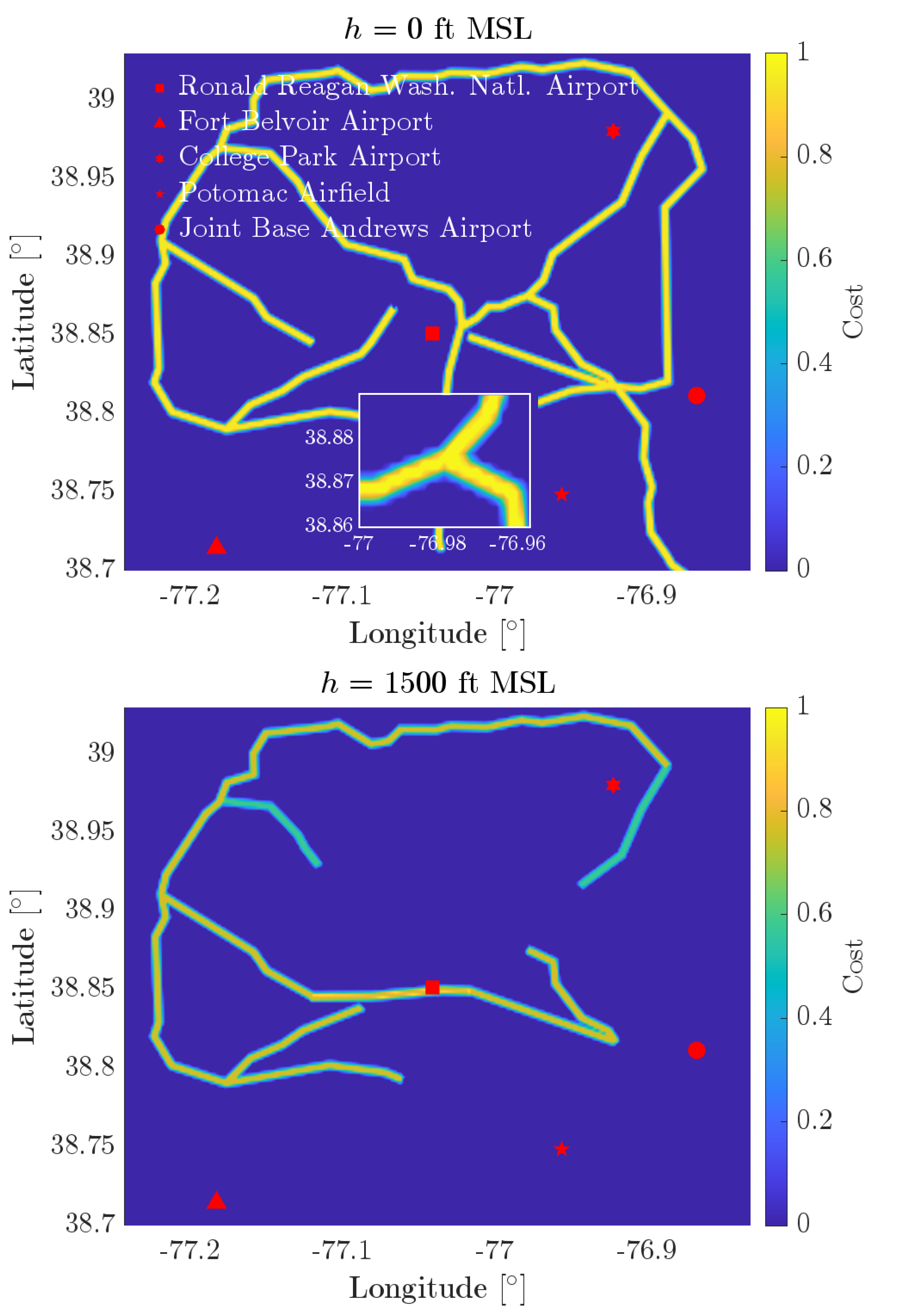}
    \caption{Heatmap of helicopter corridors in the Washington, D.C. area.}
    \label{fig:polyhedron_cost}
\end{figure}

Thus, the complete point-to-face distance evaluation depends on the angular configuration in \eqref{eq:angle_conditions}, and is computed using the orthogonal projection per \eqref{eq:point_to_plane}, edge-based projection per \eqref{eq:point_to_edge1} and \eqref{eq:point_to_edge2}, or vertex-based distance per \eqref{eq:point_to_vertex}, depending on the relative position of the point. The point-to-polyhedron minimum distance determination algorithm, including the point-in-polyhedron method, is given in Algorithm \ref{alg:point_to_polyhedron}.
\begin{algorithm}[t!]
\caption{Point-to-Polyhedron Minimum Distance}
\label{alg:point_to_polyhedron}
\begin{algorithmic}[1]
\Require Query point $\mathbf{p}$, polyhedron $P = \{f_i \mid i = 1, \ldots, 6\}$ with face normals $\vec{n}_i$ and vertices $\{\mathbf{p}_{i,j}\}$
\Ensure Minimum distance $d_{\min}(\mathbf{p} \mid P)$ and point inclusion $o(\mathbf{p} \mid P)$

\vspace{3pt}
\State $o(\mathbf{p}\mid P) \gets 1$ \Comment{Assume point is inside}
\For{$i = 1$ to $6$}
    \State Compute unit face normal vector per \eqref{eq:face_normal}
    \State Compute face centroid per \eqref{eq:centroid}
    \State $\vec{u}_i \gets \mathbf{p} - \mathbf{c}_i$
    \If{$\vec{n}_i \cdot \vec{u}_i > 0$}
        \State $o(\mathbf{p}\mid P) \gets 0$
        \State \textbf{break}
    \EndIf
\EndFor

\vspace{3pt}
\If{$o(\mathbf{p}\mid P) = 1$}
    \State \Return $d_{\min} \gets 0$, $o(\mathbf{p} \mid P)= 1$
\EndIf

\vspace{3pt}
\State $d_{\min} \gets \infty$
\For{$i = 1$ to $6$}
    \vspace{3pt}
    \State Compute angles per equation \ref{eq:angle_conditions}
    \vspace{3pt}
    \If{$\theta_1 \leq \frac{\pi}{2}$ $\land$ $\theta_2 \geq \frac{\pi}{2}$ $\land$ $\theta_3 \leq \frac{\pi}{2}$ $\land$ $\theta_4 \geq \frac{\pi}{2}$}
        \State Compute face normal $\vec{n}_i$
        \State $d \gets \lvert \vec{n}_i \cdot \mathbf{p} + k_i \rvert$ \Comment{Equation~\eqref{eq:point_to_plane}}
        \State $d_{\min} \gets \min(d_{\min}, d)$
        \State \textbf{continue}
    \EndIf

    \vspace{3pt}
    \State Define edge vectors: $\vec{v}_1 = \vec{e}_{41}$, $\vec{v}_2 = \vec{e}_{21}$
    \State Compute minimum distance with Eq. \ref{eq:point_to_edge1}, \ref{eq:point_to_edge2}, or \ref{eq:point_to_vertex}
    \State $d_{\min} \gets \min(d_{\min}, d)$
\EndFor

\vspace{3pt}
\State \Return $d_{\min}$, $o(\mathbf{p} \mid P) = 0$
\end{algorithmic}
\end{algorithm}
The cost associated with $\mathbf{p}$ relative to polyhedron $P$ is computed using $\xi:\mathbb{R}^3\times \mathbb{P} \rightarrow [0,1]$ per Eq. \ref{eq:dmin_cost}, where $d_{\max} \in \mathbb{R}^+$ denotes the maximum distance subject to penalization:
\begin{equation}
    \xi(\mathbf{p} \mid P) = 1 - \frac{\min\left(d_{\min}(\mathbf{p} \mid P),\, d_{\max}\right)}{d_{\max}}.
    \label{eq:dmin_cost}
\end{equation}
This formulation assigns a maximum penalty of one when the point is positioned inside or on one of the faces of polyhedron, and linearly decays the penalty with increasing distance up to $d_{\max}$, beyond which no cost is assigned. The cumulative cost of $\mathbf{p}$ with respect to a set of polyhedra $\mathbb{P}$ is then defined as the maximum individual penalty among all polyhedra where $\Xi:\mathbb{R}^3\times \mathbb{P} \rightarrow [0,1]$:
\begin{equation}
    \Xi(\mathbf{p},\, \mathbb{P}) = \sup_{P \in \mathbb{P}}\, \xi(\mathbf{p} \mid P).
    \label{eq:polyhedra_cost}
\end{equation}
This formulation ensures the highest risk (i.e., closest proximity to any polyhedron in $\mathbb{P}$) dominates the final cost assigned to $\mathbf{p}$.
The implementation of Algorithm \ref{alg:point_to_polyhedron} is an exhaustive search over all polyhedra. Despite its low computational overhead due to the low number of polyhedra and straightforward vector operations, heatmaps are static thus can be computed offline and loaded into the flight management system (FMS) for use in risk-aware flight planning when needed. Figure \ref{fig:polyhedron_cost} shows the risk heatmap for air corridors $\mathbb{P}_u$ at $h = 0$ ft and $h = 1500$ ft MSL for $d_{\max} = 500$ ft in the Washington, D.C. area.
A similar heatmap for no-fly zones $\mathbb{P}_n$ extending from sea level to 10000 ft is also computed for contingency landing planning.
\section{Airspace-aware Contingency Landing Planning}
This section presents a utility-based landing site selection and gradient-guided search for ground and airspace risk-aware emergency landing path planning.

\subsection{Landing Site Selection}
\label{sec:landing_site_selection}
In this study, a utility-based landing site selection algorithm is used as in \cite{atkins2006emergency,tekaslan_search}. Define landing site $l$ as
\begin{equation}
\begin{aligned}
l = \{&\varphi_t, \lambda_t, z, \chi_t, d_{\ell}, d_w, \delta^{\text{com}} , \delta^{\text{mil}} \mid \varphi_t \in \Phi, \lambda_t \in \Lambda,\\
&z \in \mathbb{R}_{\geq 0}, \chi_t \in [0, 2\pi], d_{\ell}, d_w \in \mathbb{R}^+,\\
& \delta^{\text{com}}, \delta^{\text{mil} } \in \{0,1\}\},
\end{aligned}
\end{equation}
where $\varphi_t$ and $\lambda_t$ represent latitude and longitude of the runway touchdown point, $z$ is altitude, and $\chi_t$ is runway true heading such that touchdown state is $s_t = (\varphi_t, \lambda_t, z, \chi_t)$. Runway dimensions are given by length $d_{\ell}$ and width $d_w$. Binary indicators $\delta^{\text{com}}$ and $\delta^{\text{mil}}$ specify whether the site supports commercial traffic and/or is designated as a military airfield. To enable stabilized final approaches, a virtual approach fix is generated one nautical mile from runway threshold, aligned with the reciprocal of runway heading. This fix is vertically offset from the runway touchdown point based on best-glide flight path angle, ensuring that the aircraft can safely reach the runway while maintaining a stable glide path and decelerating to the best-glide airspeed for touchdown.
Approach fix position is computed using the World Geodetic System 1984 (WGS84) model \cite{wgs84} and defined as $\mathcal{W} : \mathbb{R}^+ \times \mathbb{R}^+ \times \mathcal{S} \rightarrow \Phi \times \Lambda$. Let $c: \mathcal{S} \rightarrow \Phi \times \Lambda$ be an operator that extracts the geodetic coordinates from state $s$ such that $c(s) = [\varphi \; \lambda]^{\intercal}$. In essence, $\mathcal{W}$ returns the destination coordinates given reference state, distance, and heading. Using touchdown state $s_t$, final approach distance $d_f = 1$ NM, and reciprocal heading $\lvert \chi_t - \pi \rvert_{2\pi}$, location and altitude of approach fix $s_N$ are computed as:
\begin{equation}
c(s_N) = \mathcal{W}(d_f, \lvert \chi_t - \pi \rvert_{2\pi} \mid s_t), \quad h_N = z + d_f \tan \gamma_{bg},
\label{eq:wgs84}
\end{equation}
where the operator $\mid.\mid_{\alpha}$ wraps an angle at $\alpha$. We can further define a landing site set $\mathcal{L}$ per equation \eqref{eq:landing_site_set} where $L_i \in \mathcal{L}$.
\begin{equation}
\begin{aligned}
    \mathcal{L} = \{L_i = (l, \gamma_{0\rightarrow N}, v_{\text{hw}}) \mid &i = \mathbb{Z}_{\geq 0}, l \in \mathbb{R}^8,  \gamma_{0\rightarrow N} \in \mathbb{R},\\
    &v_{\text{hw}} \in \mathbb{R}\}
    \label{eq:landing_site_set}
\end{aligned}
\end{equation}
Here, $\gamma_{0 \rightarrow N}$ is the slope between initial state $s_0$ and approach fix $s_N$ per Eq. \eqref{eq:slope} where $d_{gc}:[\Phi\times \Lambda]\times [\Phi\times \Lambda] \rightarrow \mathbb{R}_{\geq 0}$ returns great-circle distance between the coordinates.
\begin{equation}
    \gamma_{0 \rightarrow N} =\tan^{-1}\frac{h_0 - h_N}{d_{gc}(c(s_0), c(s_N))}
    \label{eq:slope}
\end{equation}
Note that the headwind component relative to the landing site is denoted by $v_{hw}$, which is negative for tailwind.

Candidate landing sites are evaluated based on reachability, wind conditions, and operational suitability. The airports considered within this study include: Ronald Reagan Washington National Airport (DCA), Joint Base Andrews Airport (ADW), Fort Belvoir/Davison AAF (DAA), College Park Airport (CGS), Freeway Airport (W00), and Potomac Airfield Airport (VKX). ADW and DAA are military fields and may be used for emergency landings without prior authorization, per Code of Federal Regulations (CFR) 32 \S855.14~\cite{cfr855}. Runway information can be obtained from aeronautical charts \cite{skyvector}.
Candidate landing sites are ranked based on utility function $U:\mathcal{L} \rightarrow \mathbb{R}$ per \eqref{eq:util}. Site $L^{*}$ with highest $U$ is chosen.
\begin{equation}
\begin{gathered}
    U(L) = w_{\gamma}U_{\gamma} + w_wU_{w} + w_{d}U_{d} + w_cU_{c} + w_mU_m\\
    L^{*} = \arg\max_{L\in \mathcal{L}} U(L)
\end{gathered}
\label{eq:util}
\end{equation}
Here, $w_{.}$ represent utility weights of 0.5, 0.1, 0.05, 0.15, and 0.15, as ordered in \eqref{eq:util}. All utility terms of Eq. \eqref{eq:util} are functions of $L$ and expressed in Eq. \eqref{eq:util_terms}.  $U_{\gamma}$ is a reachability utility that quantifies direct flight path angle from the initial state $s_0$ to the approach fix $s_N$. The larger the value of $\gamma_{0\rightarrow N}$, the greater the altitude difference between states relative to their horizontal separation. This indicates higher available potential energy, which can be used to navigate through risky areas. Relative wind suitability is measured by wind utility $U_w$ which favors headwind for slower ground speed and shorter landing ground-roll. Longer and wider runways are prioritized with a dimension utility $U_d$ for an increased safety margins. Finally, the selection of landing sites located on military fields or open to commercial air traffic is discouraged by utility terms $U_m$ and $U_c$, respectively, to avoid military operations and interference with active air traffic corridors.
\begin{equation}
\begin{gathered}
    U_{\gamma}(L) = \frac{\gamma_{0 \rightarrow N}}{\gamma_{0 \rightarrow N,\max}}, \quad U_w(L) = \frac{v_{\text{hw}}}{v_{\text{hw},\text{max}}},\\ \quad U_{d}(L) = \frac{1}{2}\left(\frac{d_{\ell}}{d_{\ell,\max}} + \frac{d_w}{d_{w,\max}}\right),\\
    U_c(L) = 1 - \delta^{\text{com}}, \quad U_m(L) = 1 - \delta^{\text{mil}}
\end{gathered}
\label{eq:util_terms}
\end{equation}
In \eqref{eq:util_terms}, each variable with a 
max subscript in the denominator denotes the supremum of the corresponding quantity over the set of landing sites; e.g., $\gamma_{0 \rightarrow N,\max} = \sup_{L \in \mathcal{L}} L{._{\gamma_{0\rightarrow N}}}$.
    
\subsection{Dense Airspace Avoidance with Gradient-Guided Search}
\label{sec:deconfliction_search}
This section introduces airspace risk quantification as an addition to the previously defined ground-risk \cite{tekaslan_search}. Aircraft altitude must be utilized in airspace risk estimation since air traffic corridor altitude varies per approach and departure routes. Navigating around risky airspace corridors necessitates significant backtrack and extensive state-space exploration. To address, this work defines airspace risk of state $s$ as $g_a(s): \mathcal{S} \rightarrow [0,1]$ per \eqref{eq:ga}. Terms of $g_a(s)$ respectively return risk values attributed to ADS-B based air traffic and polyhedra-based air traffic corridors and prohibited zones. $w_{\kappa} = 0.5, w_u = 0.25$ and $w_p = 0.25$ are associated weights.
\begin{equation}
    g_a(s) = w_{\kappa}\kappa(\pi(s)) + w_{u}\Xi(\pi(s),\mathbb{P}_u) + w_{p}\Xi(\pi(s), \mathbb{P}_n)
    \label{eq:ga}
\end{equation}
Cumulative airspace risk $g:\mathcal{S}\rightarrow \mathbb{R}_{\geq 0}$ over the landing trajectory is given for a time-indexed path $\zeta : [0, T(s)] \rightarrow \mathcal{S}$ satisfying $\zeta(T(s)) = s$:
\begin{equation}
    g(s) = g(s \mid \mathcal{P}(s_0,s)) = w_a(s)\int_{0}^{T(s)} g_a(\zeta(t))\, dt.
    \label{eq:airspace_risk}
\end{equation}
Function $g(s)$ has seconds units and quantifies airspace risk as the time exposure of the distressed aircraft to typical air traffic routes. This formulation does allow state expansion into costly airspace corridors when needed, particularly when goal state $s_N$ is in an air corridor such that $g_a(s_N) > 0$. In such cases, search may expand numerous states to find a safer alternative trajectory. To reduce overhead while still encouraging safer paths when feasible, cumulative airspace risk is adaptively scaled by factor $w_a(s):\mathcal{S}\rightarrow [0\;1]$ as defined in \eqref{eq:w_a}. The scaling factor depends on remaining descent altitude $\Delta h = h - h_N$ to $s_N$. Upper and lower thresholds are denoted by $\overline{h} = 1500 $ ft and $\underline{h} = 1000$ ft, respectively.
\begin{equation}
    w_a(s) = 
    \begin{cases}
        \dfrac{\Delta h - \underline{h}}{\overline{h} - \underline{h}}, & \underline{h} < \Delta h < \overline{h} \text{ and } g_a(s_N) > 0 \\
        0, & \Delta h \leq \underline{h} \text{ and } g_a(s_N) > 0\\
         1, & \text{otherwise} \\
    \end{cases}
    \label{eq:w_a}
\end{equation}

\begin{figure}[t!]
    \centering
    \includegraphics[width=.6\linewidth]{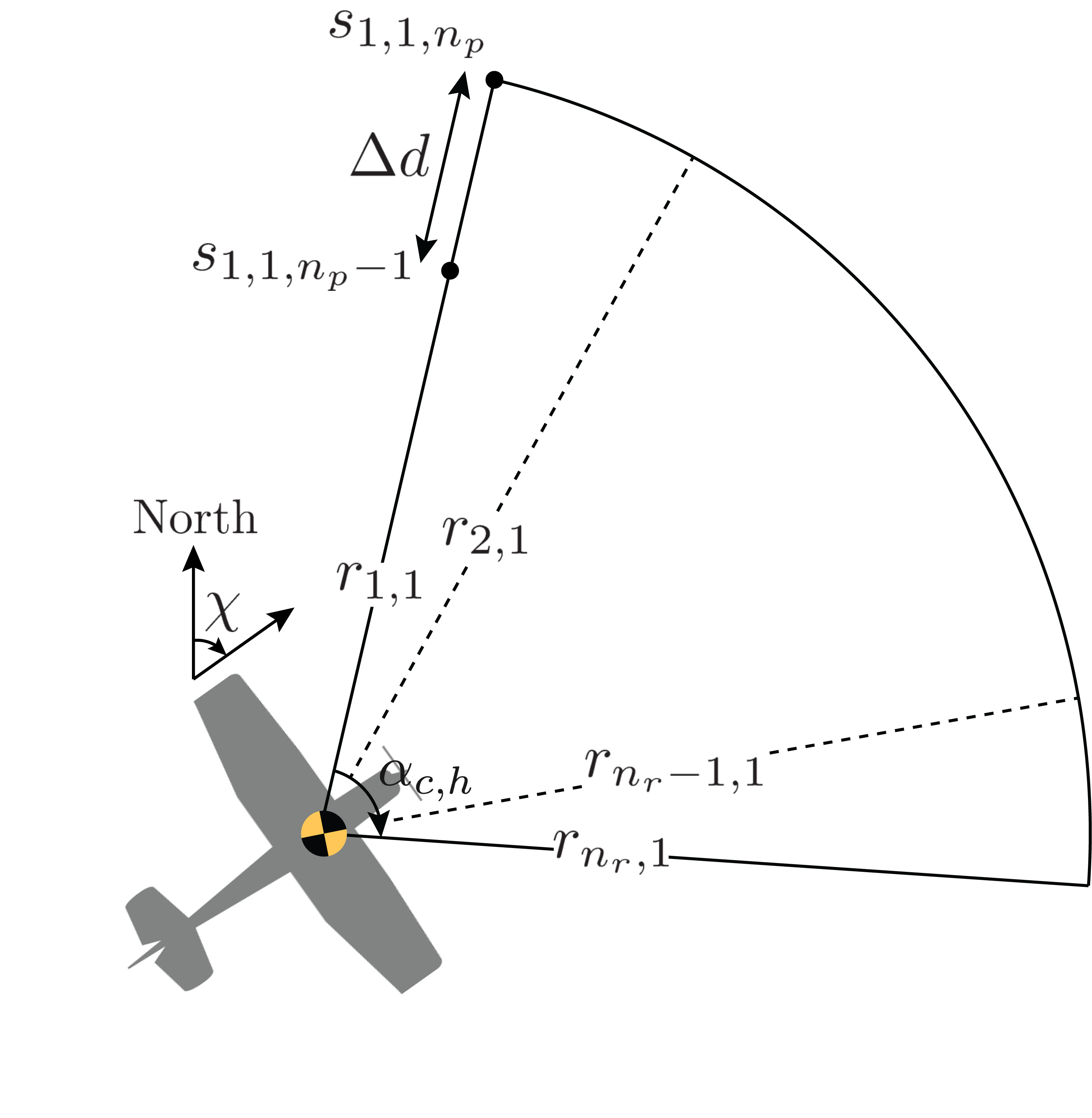}
    \caption{Airspace cost sampling in the horizontal and vertical planes.}
    \label{fig:lookahead_airpsace_cost}
\end{figure}

A look-ahead heuristic estimates future airspace risk by accounting for high-cost regions that may be encountered during subsequent state expansions. This heuristic evaluates average airspace traffic density within a cone-shaped region ahead of the aircraft by averaging airspace risk samples on cast rays as shown in Figure \ref{fig:lookahead_airpsace_cost}. This heuristic helps guide search by proactively avoiding congested airspace and improves runtime performance.
Let the airspace detection cone have angular spreads of $\alpha_{c,h}$ and $\alpha_{c,v}$ in the horizontal and vertical planes, respectively. The number of sampling rays in the horizontal and vertical directions are denoted by $n_r \in \mathbb{Z}^+$ and $n_l\in \mathbb{Z}^+$. Each ray is indexed as $r_{i,j}$, where $i \in \{1, \dots, n_r\}$ and $j \in \{1, \dots, n_l\}$ indicate the ray's orientation in the horizontal and vertical planes. The horizontal and vertical ray-casting angles, denoted by $\theta_i$ and $\theta_j$, are defined as:
\begin{equation}
    \theta_i = \chi - \frac{\alpha_{c,h}}{2} + (i-1)\frac{\alpha_{c,h}}{n_r - 1}, \quad
    \theta_j = \gamma_{bg} + (j-1)\frac{\alpha_{c,v}}{n_l - 1}.
\end{equation}
The number of state samples along each ray is denoted by $n_p\in \mathbb{Z}^+$. A sampled state is represented as $s_{i,j,k}$, where $(i,j)$ identifies the ray index and $k \in \{1, \dots, n_p\}$ is the index along the length of the ray. Each sample is separated by $\Delta d = 10\ell$ where $\ell$ is the minimum segment length defined in \cite{tekaslan_scitech}. Their positions can be computed using the WGS84 model:
\begin{equation}
\begin{gathered}
    c(s_{i,j,k}) = \mathcal{W}(k\Delta d, \theta_i \mid s)\\
    \quad h_{i,j,k} = h - k\Delta d\tan\theta_j, \quad \chi_{i,j,k} = \theta_i.
    \label{eq:airspace_sampling}
\end{gathered}
\end{equation}
The heuristic airspace risk cost ahead of the current state $s$ is computed as the average of the airspace risk over all sampled states within the forward cone, as defined in Eq. \eqref{eq:airspace_risk_heuristic}. The function $h_a(s):\mathcal{S} \rightarrow [0,1]$ is bounded within the closed interval [0,1] due to averaging the summation of normalized airspace cost terms of $g_a(s)$.
\begin{equation}
    h_a(s) = \frac{w_a(s)}{n_rn_ln_p}\sum_{i=1}^{n_r} \sum_{j=1}^{n_l} \sum_{k=1}^{n_p} g_a(\pi(s_{i,j,k}))
    \label{eq:airspace_risk_heuristic}
\end{equation}
This study uses cone angles $a_{c,h} = 60^\circ$ and $a_{c,v} = 10^\circ$ where samplings are based on $n_r = 5$, $n_l = 3$, and $n_p = 1$.

In addition, cost functions for optimal gliding, minimum remaining traversal, approach direction, and overflown population risk are used as in \cite{tekaslan_search}. Thus, total heuristic cost is:
\begin{equation}
\begin{aligned}
    h(s) &= \mathbf{w}^\intercal\mathbf{h},\\
\end{aligned}
\label{eq:total_heuristic}
\end{equation}
where weight $\mathbf{w}\in\mathbb{R}^5$ and heuristic cost $\mathbf{h}\in\mathbb{R}^5$ arrays are
\begin{equation}
    \begin{gathered}
    \mathbf{w} = 
        \begin{bmatrix}
        w_{h,1}& w_{h,2}& w_{h,3}& w_{h,4}& w_{h,5}
    \end{bmatrix}^\intercal\\
    \mathbf{h} = 
    \begin{bmatrix}
        h_{d,1}(s)& h_{d,2}(s)& h_{\chi}(s)& h_{p}(s)& h_a(s)
    \end{bmatrix}^\intercal.
    \end{gathered}
\end{equation}
The total cost of $s$ is the sum of Eqs. \eqref{eq:airspace_risk} and \eqref{eq:total_heuristic}:
\begin{equation}
    f(s) = g(s) + h(s).
\end{equation}

Cumulative airspace risk $g(s)$ tends to dominate total risk because of its magnitude. Therefore, ground risk term $h_p(s)$ in heuristic cost $h(s)$ influences state expansion only when airspace risk is negligible. Furthermore, the components of $h(s)$ may overestimate the actual cost-to-go and are therefore not admissible, with the exception of the minimum remaining traversal cost $h_{d,2}(s)$. As a result, the overall search algorithm is suboptimal. Although trials using only admissible cost term $h_{d,2}(s)$ with the airspace exposure cost $g(s)$ yield optimal solutions, often with zero airspace risk, the resulting computation times are on the order of minutes, unsuitable for real-time use. In time-critical contexts, solution availability outweighs solution optimality. 
The inadmissible heuristics may initially lead the path to remain longer in high-cost airspace. As a remedy, $h(s)$ is temporarily suppressed until a sufficient amount of altitude has been lost, allowing the planner to prioritize minimizing cumulative airspace risk in early descent. This behavior is formally defined as
\begin{equation}
   g_a(s_0) > 0 \land h_0 - h < \Delta h_{\text{cutoff}} \Rightarrow h(s) = 0,
\end{equation}
where $\Delta h_{\text{cutoff}} = 500$ ft is found suitable for allowing a C182, under optimal glide conditions, to traverse at least halfway of an airspace grid cell $C_{i,j,k}$ before heuristic guidance resumes.
\section{Use Cases and Algorithm Benchmarking}
\label{sec:results}
This section presents use cases and benchmarking of risk and computational performance in the Washington, D.C. area, highlighting how the risk-aware contingency landing planner might enhance the safety of air transportation. Similar to the approach in~\cite{tekaslan_search}, the minimum-risk Dubins solution is used for comparison and as a fallback if it outperforms the search-based planner or the search-based planner fails to return a solution within a prescribed number of state expansions.
In practice, the planner can be integrated into FMS in either a fully— or semi-autonomous mode. In the fully-autonomous case, the FMS selects the minimum-risk solution, either Dubins-based or search-based, as the emergency landing plan, even if it is slightly better. In the semi-autonomous case, the FMS presents both candidate solutions to the pilot(s) in the cockpit or to operators at a ground station, who then decide which option to execute. In either mode, the landing plan can be shared via datalink with surrounding aircraft and ATC to coordinate.

For all scenarios, a reference wind of 7 knots at 290$^\circ$ is used. Search and Dubins emergency landing trajectories are computed at the optimal flight path angles accounting for wind conditions per Eq. ~\eqref{eq:vc_opt_compact}. Two sets of 1000 emergency landing trajectories are generated for benchmarking: solutions considering airspace risk only, and solutions incorporating airspace and ground risks. The initial emergency states are quasi-randomly distributed using Halton sampling \cite{Halton1960}, which deterministically fills the unit hypercube with spaced points using prime bases. The sampled initial states lie within a bounding box. All initial samples are guaranteed reachable to at least one landing site and outside of prohibited areas beyond at least one turn radius.

Unlike the weighted ground and airspace risk cost functions used during search, the airspace risk $\sigma_a:\mathcal{S}\rightarrow \mathbb{R}_{\geq0}$ and ground risk $\sigma_g:\mathcal{S}\rightarrow \mathbb{R}_{\geq0}$ associated with a solution path $\mathcal{P}$ are computed using time integration of normalized risk values, without any weighting, for comparison purposes per \eqref{eq:risks}. The total flight time from the initial state $s_0$ to the approach fix $s_N$ is denoted by $T(s_N)$, where $s_N \in \mathcal{P}$ is the terminal state of the trajectory. Although the integrands do not explicitly reference the path, the state trajectory $\zeta(t)$ and the terminal time $T(s_N)$ are determined by $\mathcal{P}$, making the risks path-dependent.
\begin{equation}
\begin{aligned}
    \sigma_a(\mathcal{P}) = \int_{0}^{T(s_N)} g_a(\zeta(t)) dt, \quad \sigma_g(\mathcal{P}) = \int_{0}^{T(s_N)} \eta(t) dt
\end{aligned}
\label{eq:risks}
\end{equation}
Trapezoidal integration is used to compute $\sigma_a(\mathcal{P})$ and $\sigma_g(\mathcal{P})$, with time steps of 0.05 and 0.5 seconds, respectively. Joint airspace and ground risk is defined per \eqref{eq:total_risk} where $a$ is used to tune the magnitudes of risk values based on the mean values of the normalized risk distributions, $\mu_{\kappa}$ and $\mu_{\eta}$, respectively introduced in \eqref{eq:mean_kappa} and Section \ref{sec:prelim}.
\begin{equation}
    \sigma_t(\mathcal{P}) = a\sigma_a(\mathcal{P}) + (1-a)\sigma_g(\mathcal{P}), \quad a = \frac{\mu_{\eta}}{\mu_{\kappa} + \mu_{\eta}} = 0.9075
    \label{eq:total_risk}
\end{equation}

To ensure fair comparison across a wide range of risk magnitudes, a relative difference threshold is used. Specifically, for a given risk metric $\sigma \in \{\sigma_a, \sigma_g, \sigma_t\}$, the relative difference $\epsilon$ between the search-based $\mathcal{P}_s$ and Dubins-based $\mathcal{P}_d$ solutions is defined as:
\begin{equation}
    \epsilon = \frac{|\sigma(\mathcal{P}_s) - \sigma(\mathcal{P}_d)|}{\max(\sigma(\mathcal{P}_s), \sigma(\mathcal{P}_d))}.
    \label{eq:relative_error}
\end{equation}
Search and Dubins solutions are considered to have comparable risk, $\sigma(\mathcal{P}_s) \approx \sigma(\mathcal{P}_d)$, if $\epsilon \leq 0.02$ corresponding to a 2\% relative difference threshold. This formulation ensures that the comparison remains meaningful even when absolute risk values are small. For cases where $\epsilon > 0.02$, the risk is deemed significantly different, and the planner yielding the lower value is identified as the lower-risk solution. This categorization allows for assessing whether the search-based planner improves upon or underperforms the Dubins-based baseline in terms of either airspace, ground, or joint risk.

In all test cases, emergency landing planning is initially performed for the best landing site. If the search fails to return a solution within the maximum number of iterations, representing the runtime limit, it is rerun for alternate sites in descending order of utility, if available. Thus, the solution outcomes are presented in two categories: (1) best-site solutions and (2) multiple-site solutions. This distinction allows for evaluating the search-based planner’s performance under both real-time (best-site) and pre-flight (multiple-site) operational scenarios. Additional parameters for risk-aware emergency landing planning can be found in \cite{tekaslan_search}.

\subsection{Minimizing Airspace Conflict in Emergency Landing}
\label{sec:airspace_only}
In this set of benchmarking solutions, the exposure time to airspace risk, $\sigma_a(\mathcal{P})$ as defined per \eqref{eq:risks}, is evaluated for emergency landing trajectories generated by both the gradient-guided search and 3D Dubins methods. Weight array $\mathbf{w}$ used for heuristic cost calculations is:
\begin{equation}
    \mathbf{w} = 
        \begin{bmatrix}
        0.2& 0.2& 0.1& 0& 0.5
    \end{bmatrix}^\intercal.
\end{equation}

Out of 1000 total cases, the search fails to return a solution within the allowed number of state expansions in 18 cases, for which the Dubins fallback is used. The Dubins solver itself fails in seven cases, where the initial and goal states are too close to satisfy the kinematic continuity constraint and insufficient altitude prevents extension of either the initial or final approach segment. Nevertheless, at least one solution is obtained for all cases.
Among the remaining 975 converged cases, both methods yield comparable airspace risk in 259 instances with $\epsilon \leq 0.02$, typically occurring in regions with negligible air traffic, accounting for $27\%$ of the converged cases. In the remaining 716 cases, where the risk difference is significant with $\epsilon > 0.02$, the Dubins solution yields lower risk in 95 cases $\sigma_a(\mathcal{P}_s) > \sigma_a(\mathcal{P}_d)$, corresponding to $13\%$ of these cases. In the other 621 cases—accounting for $87\%$ of the cases with significant risk differences, the search-based method produces lower-risk paths, $\sigma_a(\mathcal{P}_s) < \sigma_a(\mathcal{P}_d)$.
\begin{figure}[t!]
    \centering
    \includegraphics[width=.8\linewidth]{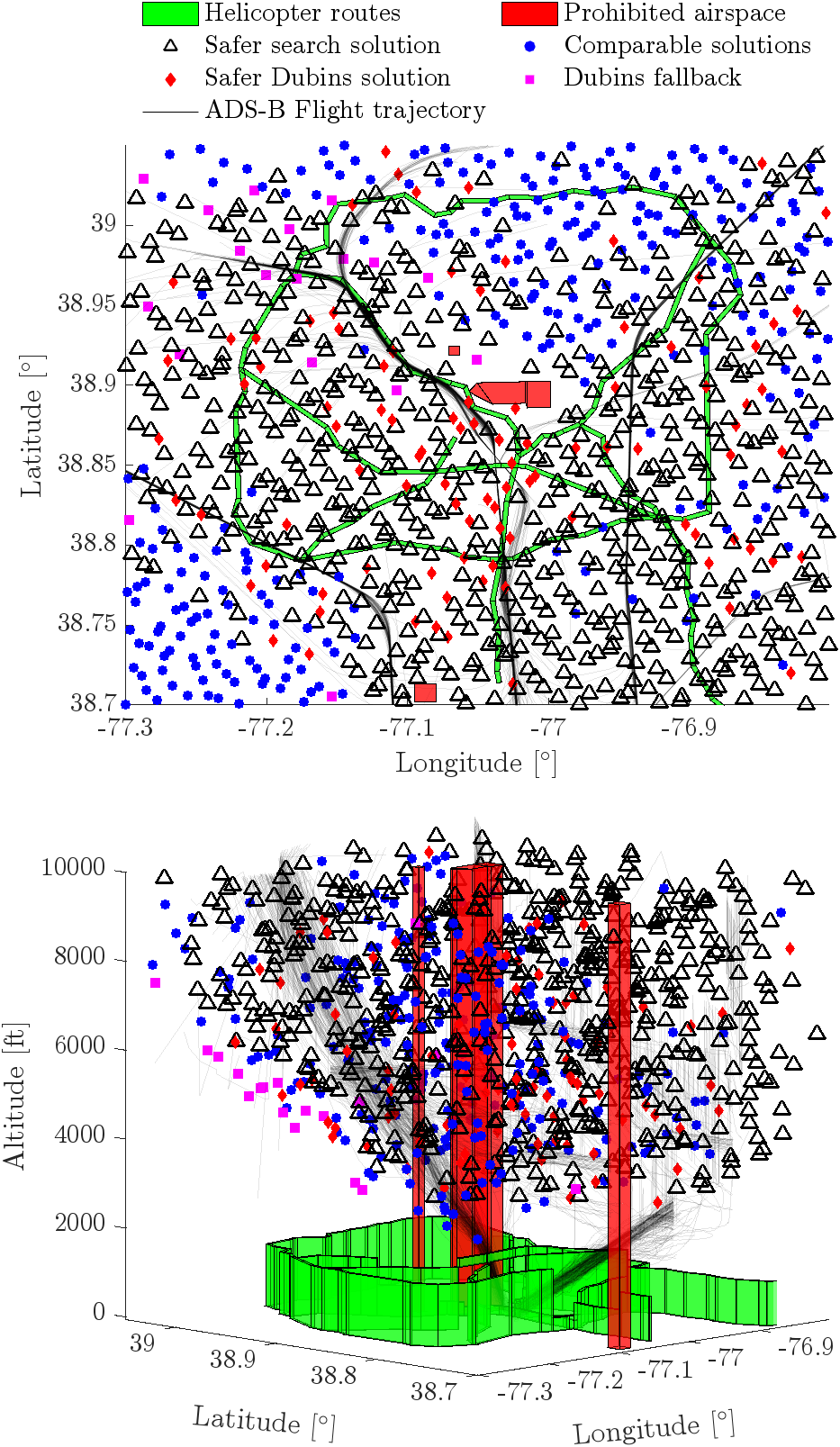}
    \caption{Randomly initialized emergency states and corresponding solution outcomes for airspace-risk minimization in emergency landing path planning.}
    \label{fig:solution_outcomes}
\end{figure}

Figure \ref{fig:solution_outcomes} illustrates randomly initialized emergency states and their corresponding landing solution outcomes. The 3D view reveals that fallback to Dubins solutions, indicated by magenta squares, forms a cone-like 3D boundary. This is expected as the cost formulation encourages using available potential energy to avoid high-risk airspace, resulting in a trade-off between reachability and exposure to air traffic corridors. Consequently, the search-based solver struggles to find feasible trajectories in marginally reachable cases, where minimum-risk Dubins solutions often lie. The asymmetrical altitude distribution of solutions found in the Northwest and Southeast is due to the exclusion of airports in the Northwest region of the map, such as Dulles International Airport.

The top view highlights clustered blue circles representing comparable-risk solutions. These cases tend to occur either in areas with minimal air traffic, such as the Southwestern map region, or when the emergency initiates at or below the altitude of typical air traffic corridors, as observed in the Northern region. For emergency states located in constrained airspace, such as around departure/arrival routes, urban air corridors, or no-fly zones, the search-based approach dominates the Dubins solver in lower-risk trajectory generation. This suggests that the search-based method proactively avoids congested airspace thus will minimize disruption to other air traffic. 

Descriptive statistics of airspace risk, shown in Table~\ref{tab:airspace_only_stats}, support the overall findings for both planning to the best landing site alone and for cases where alternate sites are considered alongside the best option. To further emphasize the differences, statistics of risk exposure to the average normalized air traffic density are also reported. Risk is quantified in seconds, corresponding to the time an aircraft would spend in regions of the airspace with a normalized cost of 1. However, the airspace as a whole has a much lower average cost of 0.014. Consequently, what appear to be small differences between the search-based and Dubins-based solutions actually correspond to substantial differences in exposure time when mapped onto the real distribution of traffic density.
\begin{table*}[t!]
    \centering
    \caption{Descriptive statistics of airspace risk (in seconds) for search-based and Dubins solutions across converged cases, under maximum ($\bar{\kappa}=1$) and average ($\bar{\kappa}=0.014$) normalized air traffic density.}
    \renewcommand{\arraystretch}{1.1}
    \begin{tabular}{lclccccccc}
    \hline \hline
    \textbf{Landing Site} & \textbf{\# Solutions} & \textbf{Density} &  \textbf{Method} & \textbf{Min.} & \textbf{Max.} & \textbf{Mean} & \textbf{Median} & \textbf{Std. Dev.} \\
    \hline
    \multirow{4}{*}{Best site} & \multirow{4}{*}{760} & \multirow{2}{*}{$\bar{\kappa} = 1$} & Search & 0 & 45.91 & 3.37 & 0.08 & 7.23\\
    & & & Dubins & 0 & 86.87 & 6.28 & 0.55 & 13.11\\
    \cline{3-9}
    & & \multirow{2}{*}{$\bar{\kappa} = 0.014$} & Search & 0 & 3278.94 & 240.91 & 5.83 & 516.71\\
    & & & Dubins & 0 & 6205.17 & 448.24 & 39.30 & 936.38\\
    \hline
    \multirow{4}{*}{Multiple site} & \multirow{4}{*}{975} & \multirow{2}{*}{$\bar{\kappa} = 1$} & Search & 0 & 49.26 & 3.93 & 0.13 & 7.76\\
    & & & Dubins & 0 & 88.93 & 8.13 & 0.738 & 15.05\\
    \cline{3-9}
    & & \multirow{2}{*}{$\bar{\kappa} = 0.014$} & Search & 0 & 3518.87 & 280.63 & 9.44 & 554.16\\
    & & & Dubins & 0 & 6352.14 & 580.83 & 52.698 & 1074.97\\
    \hline
    \end{tabular}
    \label{tab:airspace_only_stats}
\end{table*}
Overall, the search-based planner lowers both maximum and mean airspace risk exposure, enhancing the resilience of nominal operations during emergency landing planning.

\begin{figure}[t!]
    \centering
    \includegraphics[width=.9\linewidth]{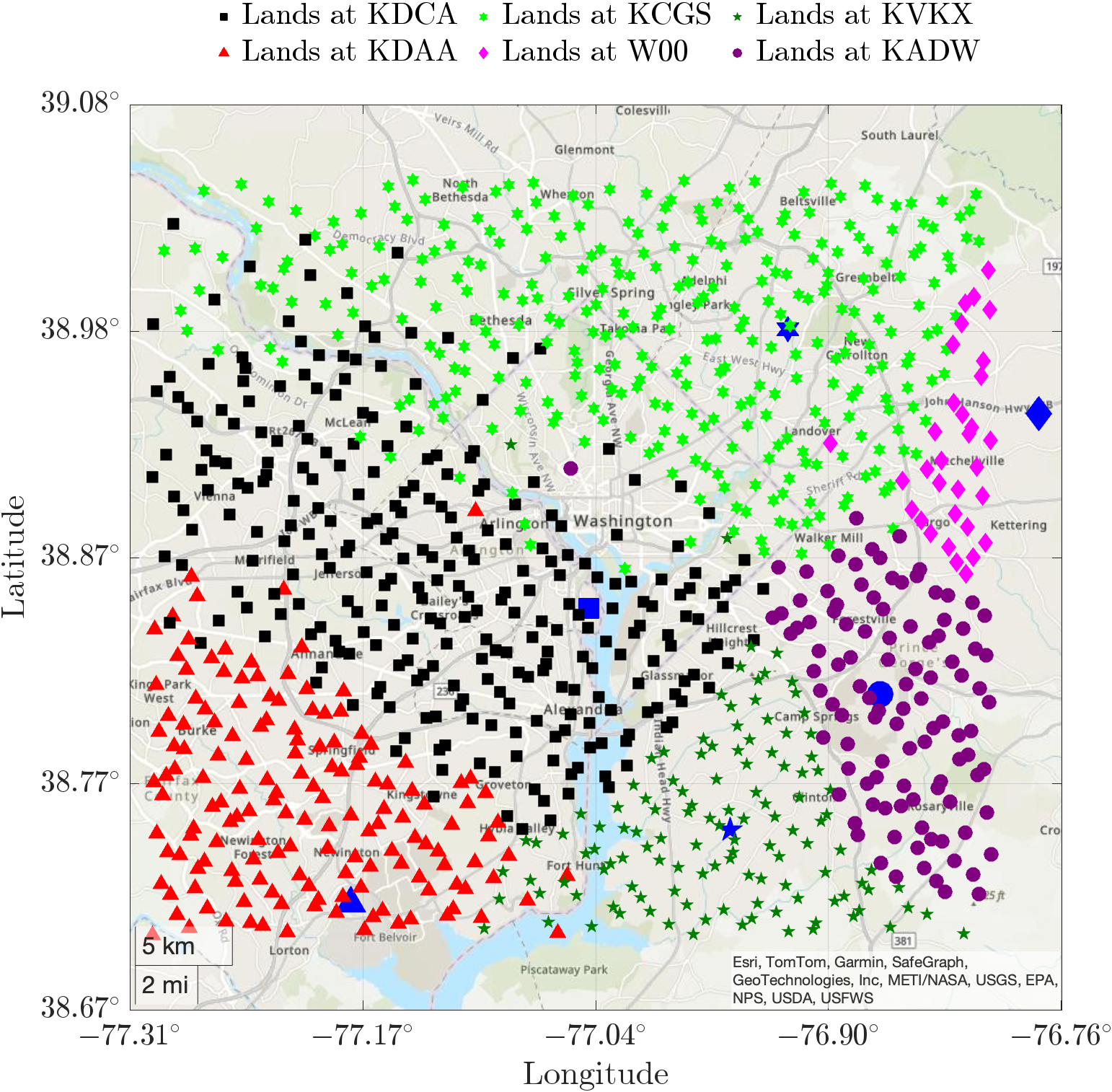}
    \caption{Initial emergency states and corresponding emergency landing sites.}
    \label{fig:landing_site}
\end{figure}

Figure \ref{fig:landing_site} shows initial emergency states and highest utility airports chosen for emergency landings.
Out of 975 converged cases, landing trajectories are generated to the best landing site in 760 cases. In the remaining cases, alternate runways or airports are used when search fails to return a solution to the best site within the allowed iterations. As the runtime constraint is relaxed by increasing the maximum number of state expansions, a path to the highest utility landing site is ultimately found due to the convergence guarantee of search-based planning \cite{tekaslan_search}. Table \ref{tab:landing_site_stats} summarizes the frequency with which each airport is selected as the landing site across all cases. Overall, College Park Airport and Ronald Reagan Washington National Airport are most frequently selected. College Park is preferred because it does not offer commercial operations and is not a military airfield, while DCA is centrally located on the map, offering higher reachability. Note that the number of landings for specified airports is expected to change depending on the geodetic bounding box.
\begin{table}[t!]
    \centering
    \caption{Landing sites across all cases.}
    \begin{tabular}{lc}
    \hline \hline
    \textbf{Airport} & \textbf{Number of landings}\\
    \hline
     Ronald Reagan Wash. Natl. Airport & 278\\
     Fort Belvoir Airport & 151\\
     College Park Airport & 314\\
     Freeway Airport & 32\\
     Potomac Airfield & 121\\
     Joint Base Andrews Airport & 104\\
     \hline
    \end{tabular}
    \label{tab:landing_site_stats}
\end{table}

\begin{figure*}[t!]
    \centering
        \centering
        \includegraphics[width=0.9\linewidth]{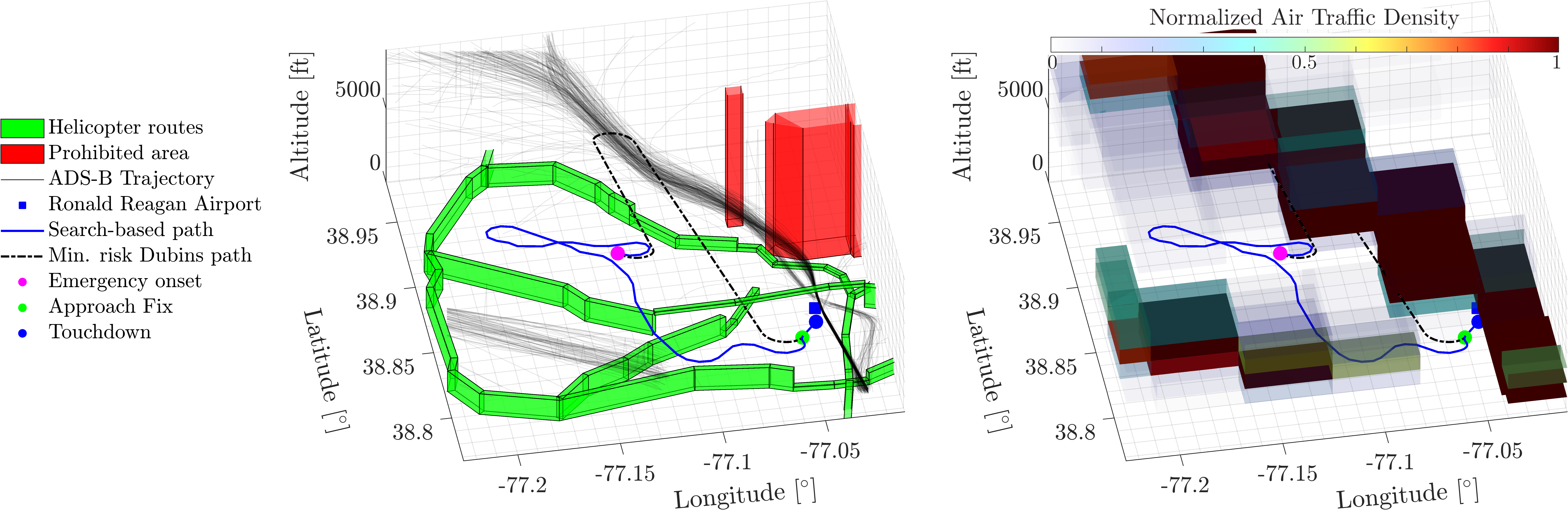}
        \caption{Use case with the maximum airspace risk difference between search- and Dubins-based solutions with risk values of 0.04 s and 79.8 s, respectively. \textbf{Left:} 3D view with the helicopter routes, no-fly zones, and ADS-B trajectories. \textbf{Right:} 3D view with the normalized air traffic density grid.}
        \label{fig:best_case}
\end{figure*}

The first use case solution in which the airspace risk difference is the absolute maximum within the benchmarking solution set in favor of search-based planner is exhibited in Figure \ref{fig:best_case}. The emergency onset is situated 4.9 NM West of Ronald Reagan Washington National Airport at 8309 ft MSL, between departure and arrival corridors. Since the descent angle $\gamma_{0\rightarrow N}$ is steep relative to the landing site, Runway 04 at DCA, excess altitude must be dissipated before the final approach. The Dubins solver accomplishes this by extending the path northward. Although the S-turn maneuver avoids the elevated airspace risk posed by arriving traffic on the downwind and final approach paths to the South, it still transits through departure traffic, resulting in a high airspace risk exposure of $\sigma_{a}(\mathcal{P}_d) = 79.8$ seconds. Per the integration in~\eqref{eq:risks}, this risk value corresponds to spending 79.8 seconds in airspace with a normalized risk of $\bar{\kappa} = 1$, or equivalently 159.6 seconds in airspace with half the risk, $\bar{\kappa} = 0.5$.

In contrast, the search-based planner navigates between dense departure and arrival corridors, utilizing a traffic-free gap where it temporarily loiters by following the gradient field of $h(s)$. As the trajectory converges toward the optimal glide altitude for the remaining traversal, it is guided toward the designated approach fix, yielding an airspace risk exposure of $\sigma_a(\mathcal{P}_s) = 0.04$ seconds—equivalent to being exposed to airspace with normalized cost $\bar{\kappa} = 1$ for 0.04 seconds.

Solutions for the second case where airspace risk difference is greatest favor the Dubins-based planner as shown in Figure \ref{fig:best_case}. Similar to the first use case, the emergency is initialized near DCA specifically 3.9 NM to the South of the field at 5589 ft MSL. Since the landing is planned to Runway 01, actively used for departures and arrivals based on the ADS-B data, significant disruption to normal air traffic operations is anticipated for both solutions. That said, the search-based landing trajectory remains in the arrival corridor for a longer duration in this particular case. This underperformance stems from the altitude weight $w_a(s)$, which improves convergence in high-density grids but reduces the influence of airspace risk as the expansion approaches the goal state. Note that once the aircraft is committed to landing, its altitude and therefore its reachability are low, leaving no alternative landing sites. In such emergencies, airport traffic would be halted to ensure no other aircraft conflict with the distressed aircraft.
\begin{figure*}[t!]
        \centering
        \includegraphics[width=0.9\linewidth]{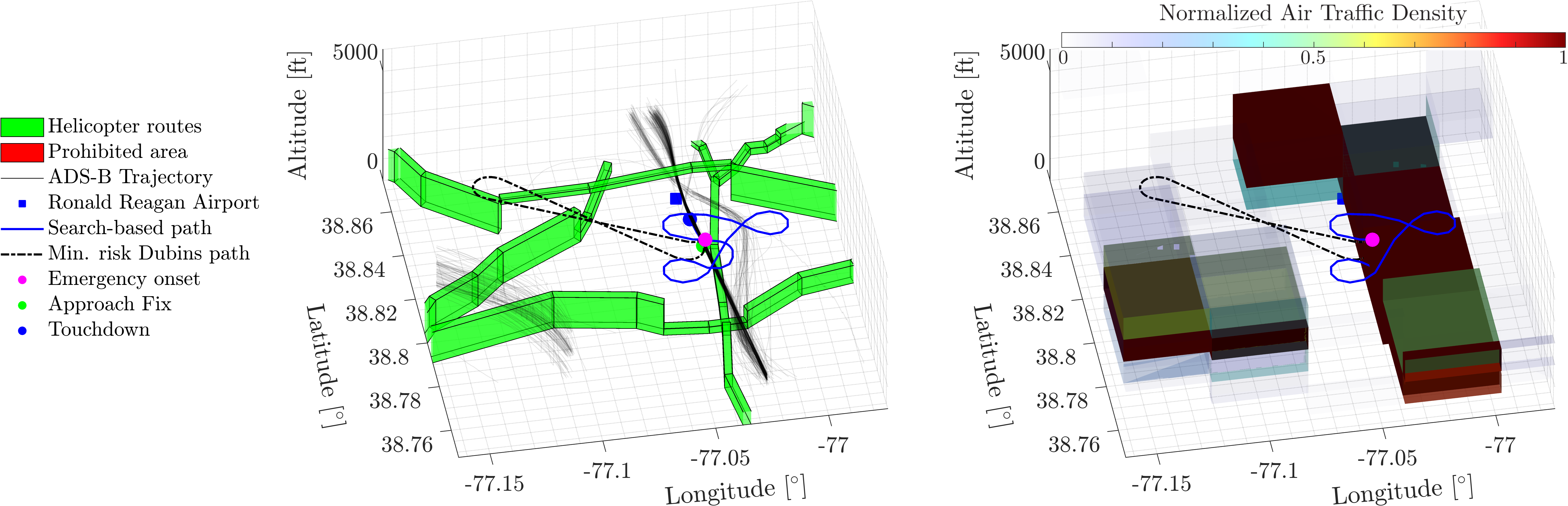}
        \caption{Use case with the minimum airspace risk difference between search- and Dubins-based solutions with risk values of 39.6 s and 11.8 s, respectively. \textbf{Left:} 3D view with the helicopter routes, no-fly zones, and ADS-B trajectories. \textbf{Right:} 3D view with the normalized air traffic density grid.}
        \label{fig:worst_case}
\end{figure*}

\subsection{Joint Minimization of Airspace Conflict and Ground Risk}
The ground risk $\sigma_g(\mathcal{P})$ expresses a normalized measure of human and property risk exposure due to the aircraft's path during an emergency descent. It quantifies how densely populated the areas are under the emergency flight path over time. In this set of benchmarking solutions, how effective the total airspace and ground risk minimization of both methods are investigated. Following \cite{tekaslan_search}, the ground-risk term is ignored above a crossover altitude of 5000 ft MSL, where the landing trajectory imposes minimal ground exposure. The total joint risk is a linear combination of risk factors, scaled based on the average risk within the considered airspace per~\eqref{eq:total_risk}. Weight array $\mathbf{w}$ used for heuristic cost calculations is:
\begin{equation}
    \mathbf{w} = 
        \begin{bmatrix}
        0.1& 0.1& 0.05& 0.5 & 0.25
    \end{bmatrix}^\intercal.
\end{equation}

\begin{figure}[t!]
    \centering
    \includegraphics[width=0.9\linewidth]{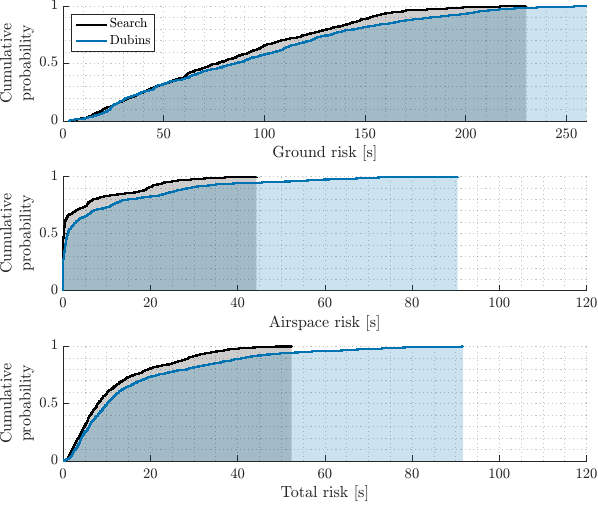}
    \caption{Airspace and ground risk avoidance cumulative probability distributions for search and Dubins emergency landing path planners.}
    \label{fig:risk_comparison_cumul_dist}
\end{figure}

Among 1000 cases, the search-based planner fails to converge in 30 instances, whereas the Dubins approach yields no kinematically feasible solution in seven cases as previously. Accordingly, the comparative analysis is conducted among the 963 cases in which both methods produce converged solutions. In terms of ground risk mitigation performance, both methods yield closely matched results in 37 cases with $\epsilon \leq 0.02$, accounting for $4\%$ of the cases. While the search-based planner achieves lower ground risk in 559 instances, Dubins outperforms search in 367 cases. Thus, search yields better solutions in $60\%$ of the cases where risk difference is significant.
Regarding airspace risk, similar performance is observed in 240 cases, while Dubins performs better in 78, and the search-based planner outperforms in 645 cases—$89\%$ of the remaining cases. Regarding the total risk per \eqref{eq:total_risk}, search achieves better results than Dubins in 697 cases, corresponding to $72\%$ success. Figure \ref{fig:risk_comparison_cumul_dist} exhibits the comparison of cumulative probability distributions of ground, airspace, and combined risks obtained with both methods.
As observed, the search-based method has a higher probability of returning an emergency landing solution with total risk less than or equal to an arbitrary threshold. Total risk statistics are detailed in Table \ref{tab:airspace_ground_stats}. Similar to the individual risk performance discussed earlier, the search-based planner significantly improves the worst-case scenario compared to the Dubins solver.
\begin{table*}[t!]
    \centering
    \caption{Descriptive statistics of joint airspace and ground risk for search-based and Dubins solutions across converged cases.}
    \renewcommand{\arraystretch}{1.1}
    \begin{tabular}{lcccccccc}
    \hline \hline
    \textbf{Landing Site} & \textbf{\# Solutions} & \textbf{Method} & \textbf{Min.} & \textbf{Max.} & \textbf{Mean} & \textbf{Median} & \textbf{Std. Dev.} \\
    \hline
    \multirow{2}{*}{Best site} & \multirow{2}{*}{730} & Search & 0.38 & 52.38 & 10.28 & 6.88 & 10.04\\
    & & Dubins & 0.27 & 91.70 & 13.72 & 8.39 & 14.92\\
    \hline
    \multirow{2}{*}{Multiple sites} & \multirow{2}{*}{963} & Search & 0.38 & 52.38 & 11.86 & 8.00 & 10.53\\
    & & Dubins & 0.27 & 91.70 & 16.82 & 10.154 & 17.46\\
    \hline
    \end{tabular}
    \label{tab:airspace_ground_stats}
\end{table*}

\begin{figure*}[t!]
    \centering
    \includegraphics[width=\linewidth]{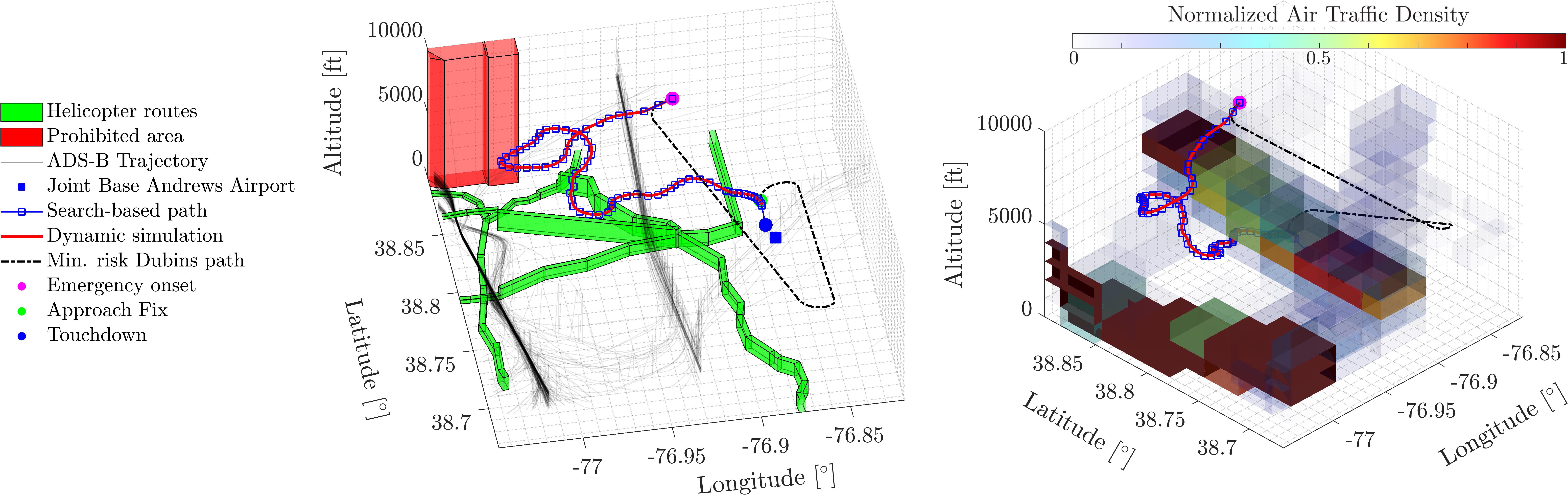}
    \caption{Dynamic simulation of the landing plan for the use case with the maximum ground risk difference between search-based and Dubins-based solutions. Search ground risk = 148 s, Dubins ground risk = 27.6 s, Search airspace risk: 0.32 s, Dubins airspace risk: 5.6 s.}
    \label{fig:use_case_joint}
\end{figure*}

\begin{figure}[t!]
    \centering
    \includegraphics[width=0.9\linewidth]{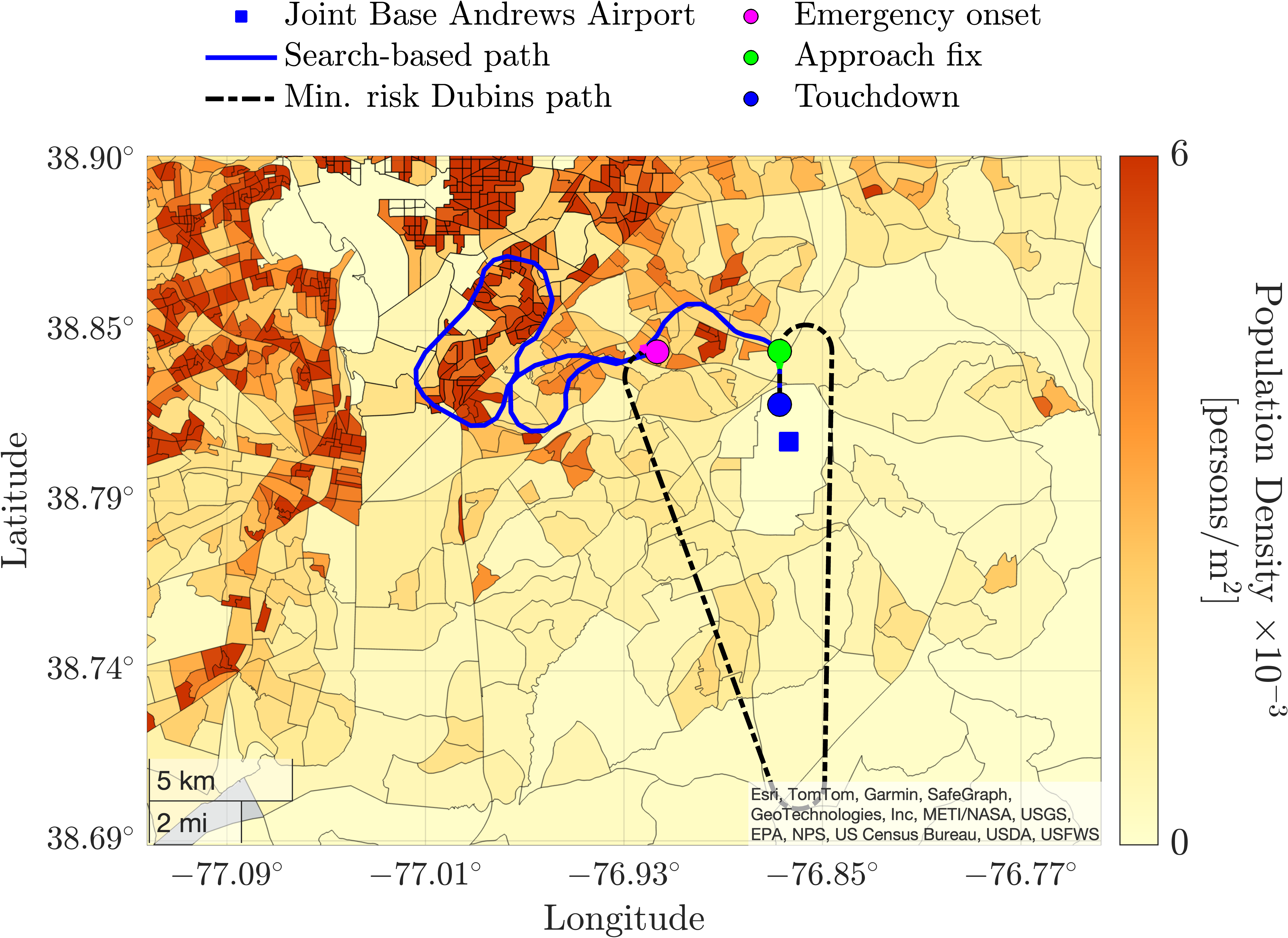}
    \caption{Joint risk minimization use case solution on a population map.}
    \label{fig:use_case_joint_pop}
\end{figure}

Figures~\ref{fig:use_case_joint} and ~\ref{fig:use_case_joint_pop} show joint risk mitigation use case results, respectively depicting the 3D trajectory with airspace risk factors and the top-down view over a population density map. The emergency is initialized 3 NM Northwest of Joint Base Andrews Airport at an altitude of 9429 ft MSL. This particular case represents the largest absolute difference in ground risk across the solution set, favoring the Dubins-based solver: $\sigma_g(\mathcal{P}_d) = 27.6$ s versus $\sigma_g(\mathcal{P}_s) = 148$ s. However, the search-based solution yields lower airspace risk: $\sigma_a(\mathcal{P}_d) = 0.32$ s and $\sigma_a(\mathcal{P}_s) = 5.6$ s. The higher airspace risk in the Dubins-based trajectory arises from its southward extension, which avoids arriving traffic on downwind of Ronald Reagan Airport. This detour intersects less dense arrival corridor over sparsely populated areas, resulting in a lower cumulative ground risk as shown in Figure~\ref{fig:use_case_joint_pop}. On the other hand, the search-based planner explicitly avoids air traffic, reducing airspace disruptions at the expense of traversing more densely populated areas and thereby increasing ground risk. Because emergency onset occurs 429 ft above an air corridor, expansion begins westward; an eastward expansion with a turn would yield comparable traffic exposure, whereas the look-ahead heuristic $h_a(s)$ discourages North and South moves due to the corridor directly beneath the initial position. In addition, the optimal glide $h_{d,1}(s)$ and minimum remaining traversal $h_{d,2}(s)$ heuristics limit excursions far from the landing site in contrast to the Dubins solver’s S-turn tendency. This case exemplifies the trade-off between airspace and ground risks encoded in the cumulative risk objective and associated heuristics.

\subsection{Dynamic Feasibility Validation}
\begin{figure}[t]
    \centering
    \includegraphics[width=\linewidth]{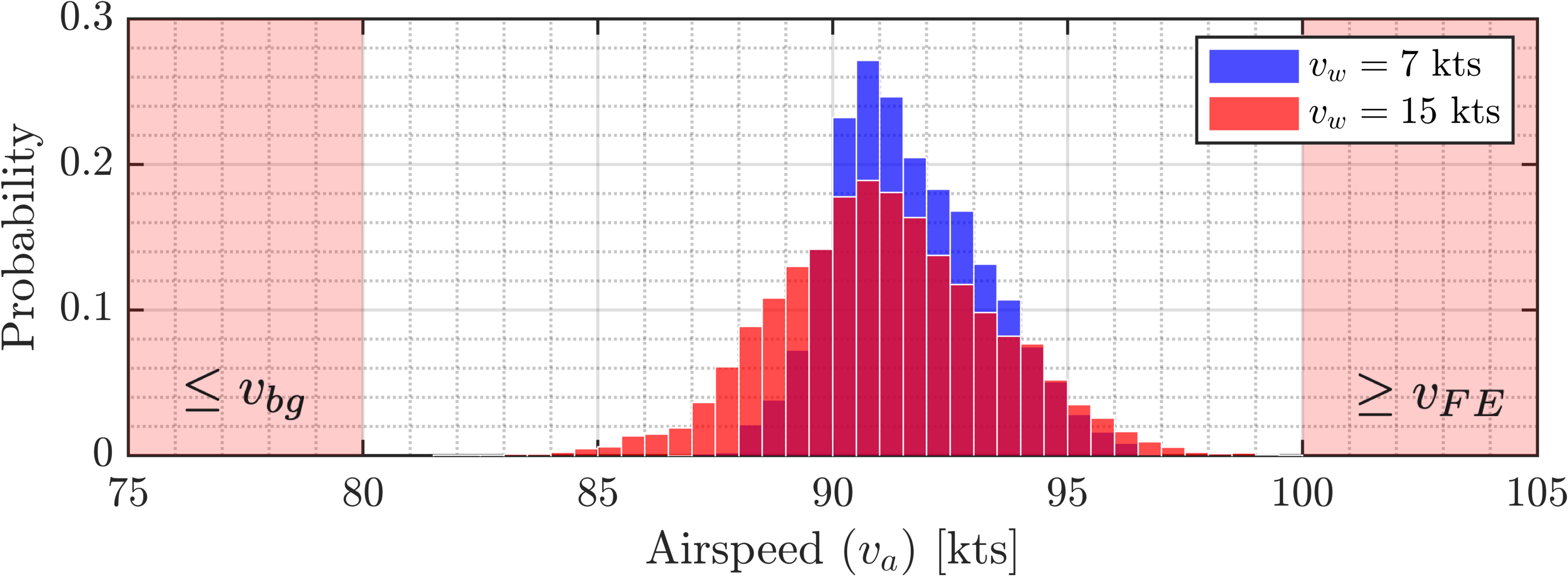}
    \caption{Airspeed distributions for dynamic landing simulations.}
    \label{fig:vadist}
\end{figure}

Dynamic feasibility is validated by executing search-based landing paths in a six-degree-of-freedom engine-out C182 simulation \cite{tekaslan2025_v2v}. The model employs a proportional–integral–derivative attitude controller, fly-by waypoint guidance to track search-generated segments, and a Dubins guidance mode for final runway alignment. Figure~\ref{fig:use_case_joint} illustrates a simulated landing trajectory (red); small lateral and longitudinal tracking errors indicate that the action set is dynamically feasible. To assess robustness, 200 airspace-risk-only cases are randomly selected and the same set is executed under two steady wind conditions: 7 knots from 290$^\circ$ and 15 knots from 290$^\circ$, matching the C182’s maximum demonstrated crosswind for takeoff and landing \cite{c182poh}. Figure~\ref{fig:vadist} presents airspeed distributions, with detailed statistics reported in Table~\ref{tab:vaStats}. Results show the open-loop airspeed is always maintained within the acceptable margin even in strong winds.
\begin{table}[t!]
    \centering
    \caption{Airspeed statistics from dynamic simulations in knots.}
    \renewcommand{\arraystretch}{1.1}
    \begin{tabular}{lcccc}
     \hline \hline
     \textbf{Wind} & \textbf{Min.} & \textbf{Max.} & \textbf{Mean}& \textbf{Std. Dev.}\\
     \hline
     7 knots & 87 & 96.7 & 91.7 & 1.6\\
     15 knots & 81.6 & 99.7 & 91.1 & 2.3\\
    \hline
    \end{tabular}
    \label{tab:vaStats}
\end{table}

\subsection{Runtime Performance Analysis}
\begin{figure}
    \centering
    \includegraphics[width=0.9\linewidth]{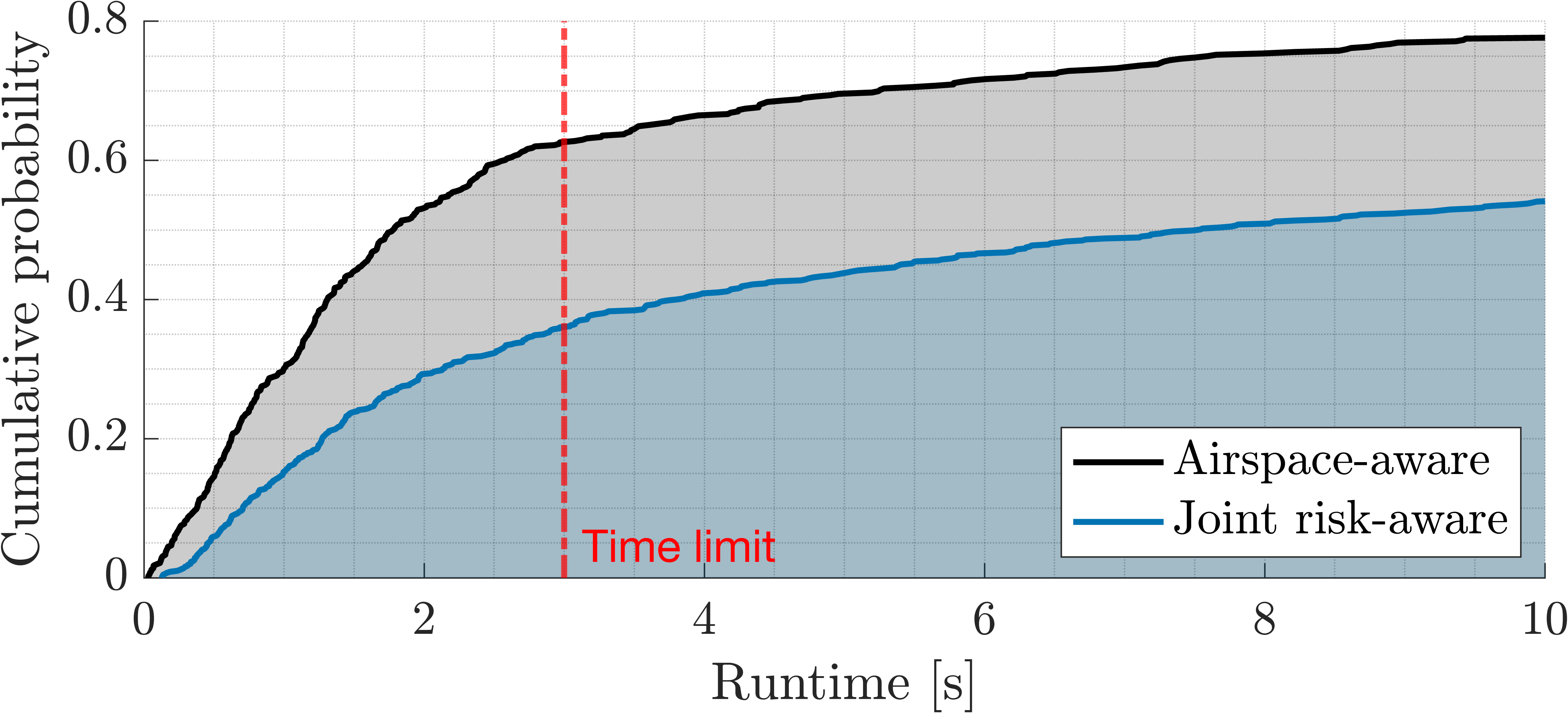}
    \caption{Cumulative probability distribution of emergency landing path planning runtime and a vertical red line indicating the prescribed time limit.}
    \label{fig:runtime}
\end{figure}

The search-based planner is programmed in C/C++. Runtime results presented here are obtained on a personal laptop with Apple M2 chip. Pitch attitude stabilization required approximately three seconds for the Cessna 182 following an engine failure in cruise flight \cite{tekaslan2025_v2v} thus is set as the time limit for real-time emergency landing path planning. Let $t$ be the online runtime, and $t_{\text{max}}$ be the time limit posed by the pre-defined maximum number of state expansions for the sake of algorithm overhead analysis. To accurately assess the real-time performance of the search-based planner, only solutions that result in landing at the best site are considered. In these cases, the planner is executed only once, and the recorded runtime directly reflects its convergence behavior. In contrast, alternate-site landings typically involve multiple sequential planning attempts, introducing overhead that obscures the true runtime of a single search execution. Therefore, the joint cumulative distribution of runtime and search-based risk improvement is estimated, conditioned on cases where the landing is to the best site, to ensure a faithful assessment of real-time performance. Let $E_1$ be the event of search significantly improving the risk across all events $E$—all converged solutions:
\begin{equation}
    E_1 \subset E = \sigma(\mathcal{P}_s) < \sigma(\mathcal{P}_d)\;\cap\; \epsilon > 0.02.
\end{equation}
Let $E_2$ be best-site landing cases with large risk difference:
\begin{equation}
    E_2 \subset E = \text{Best-site landings}\; \cap \; \epsilon > 0.02.
\end{equation}
Cumulative probability of search executing to completion by time $t$, conditioned on events $E_1$ and $E_2$, can be expressed using a conditional cumulative distribution function of runtime:
\begin{equation}
    F_T(t \mid E_{1}, E_{2}) = 
    \int_{0}^{t_{\max}} f_T(t \mid E_1, E_2)\, dt.
    \label{eq:conditional_cdf}
\end{equation}
Here, $f_T(t \mid E_1, E_2)$ denotes the conditional probability density function of runtime, given both events. To reflect the overall cumulative probability, $f_T(t\mid E_1, E_2)$ is integrated in time and scaled by the empirical probability $\text{Pr}(E_1 \mid E_2)$. This leads to the following expression:
\begin{equation}
    F(t, E_1 \mid E_2) = 
    \Pr(E_1 \mid E_2) \cdot 
    F_T(t \mid E_{1}, E_{2}).
    \label{eq:scaled_cdf}
\end{equation}
Equation~\eqref{eq:scaled_cdf} represents the cumulative probability that the planner converges within time $t$ given that search executes to completion and yields an improvement over the Dubins baseline. Figure \ref{fig:runtime} visualizes \eqref{eq:scaled_cdf} based on benchmarking observations. 
Within prescribed planning time limit $t_{\max} = 3$ sec, search-based planning yields superior airspace risk mitigation compared to Dubins with a probability of $62\%$. In other words, search either exceeds $t_{\max}$ and/or it underperforms relative to Dubins in the remaining $38\%$ of all cases. The cumulative probability of obtaining a solution that improves the joint risk metric relative to Dubins under the same limit is $36\%$, highlighting the reduced real-time performance resulting from ground risk evaluation. For larger values of $t_{\max}$, as may occur in less urgent or higher-altitude emergencies, convergence rates increase for individual and joint risk metrics.

A summary of the runtime statistics is presented in Table~\ref{tab:runtime}. For airspace risk avoidance, the search-based planner achieves an average convergence time of approximately 2.9 seconds. It is well within the disruption window reported by the European Aviation Safety Agency Startle Effect Management report for pilots \cite{EASA2018Startle}, where human startle can impair complex motor tasks such as decision-making and checklist execution for up to 10 seconds, thereby demonstrating feasibility for real-time autonomous emergency landing path planning.
When ground risk is incorporated into a joint risk metric, the average computation time increases to about 9 seconds, indicating higher computational complexity but still within a range that may be suitable for urgent in-flight planning depending on system capabilities.
\begin{table*}[t!]
    \centering
    \caption{Summary runtime statistics of search-based emergency landing path planning in milliseconds.}
    \renewcommand{\arraystretch}{1.1}
    \begin{tabular}{llccccc}
     \hline \hline
     \textbf{Risk Metric} & \textbf{Landing Site} & \textbf{Min.} & \textbf{Max.} & \textbf{Mean} & \textbf{Median} & \textbf{Std. Dev.}\\
     \hline
     \multirow{2}{*}{Airspace} & Best site & 24 & 27668 & 2895 & 1336 & 4650\\
     & Multiple sites & 20 & 263513 & 15751 & 1940 & 32387\\
    \hline
    \multirow{2}{*}{Joint} & Best site & 25 & 54715 & 9020 & 3862 & 11164\\
     & Multiple sites & 25 & 366465 & 29673 & 7957 & 50602\\
    \hline
    \end{tabular}
    \label{tab:runtime}
\end{table*}
Considering multiple landing sites, the average planning runtime increases considerably to 15.7 seconds for airspace risk and nearly 30 seconds for joint risk, reflecting the additional overhead of evaluating multiple options. These longer runtime suggest that the planner is better suited for offline, pre-flight contingency management when generating multiple candidate trajectories in advance. In this mode, the flight management system can preemptively compute emergency landing paths for a range of aircraft states on a flight plan prior to flight, allowing low-latency retrieval of risk-aware landing trajectories during emergencies.
\section{Discussion}
\label{sec:discussion}
This work utilized a historical ADS-B dataset to estimate air traffic density. For practical applications, the dataset 
must be indexed and updated to accommodate protocol updates and factors such as weather conditions, bird activity, or noise abatement procedures. Pre-flight updates can be retrieved using filed flight plans, airport operational data, and Meteorological Aerodrome Reports (METARs). Additionally, the dataset can be updated in-flight via satellite datalink \cite{8935306} to account for operational changes. The planner may execute on a ground station, enabling real-time solutions to be generated and uplinked to nearby traffic for coordinated emergency response.

\begin{figure}[t!]
    \centering

\includegraphics[width=\linewidth]{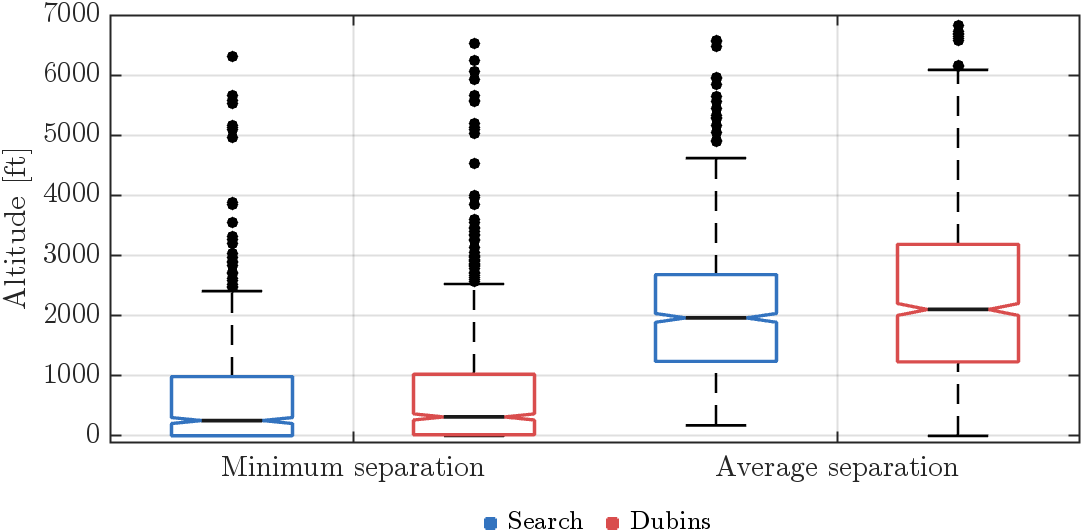}
    \caption{Vertical separation statistics for search- and Dubins-based solutions.}
    \label{fig:separation}
\end{figure}

Use case results show search-based trajectories can weave above or below high-cost air traffic corridors. Figure \ref{fig:separation} reports vertical separation statistics for search and Dubins solutions, both of which may still produce air corridor crossings. Search may follow corridor surfaces because proximity is not penalized. 
To complement the presented risk mitigation strategy, an additional tool is envisioned to identify air traffic that may conflict with the shared emergency landing plan and to adjust surrounding trajectories accordingly. This coordination must assure  potential conflicts are resolved proactively, minimizing the overall impact on ongoing air traffic operations.

Results indicate the Dubins solver can also produce low-risk trajectories, making it a valuable complementary approach. Additionally, search may fail to return a solution within runtime limits, whereas the Dubins solver always generates feasible paths in well under a second. The search-based planner exhibits reduced performance in mitigating ground risk than airspace risk, largely due to the structure of its cost formulation. This can be addressed by incorporating cumulative ground risk directly into the cost function—analogous to the treatment of airspace risk—rather than relying on a heuristic term. However, such a modification would increase state-space exploration and in turn computation overhead.

\section{Conclusion}
\label{sec:conclusion}
This paper presents an airspace risk–aware emergency landing path planning algorithm that supports deconfliction in dense airspace environments for fixed-wing aircraft. Risk is defined as the time an emergency trajectory spends within densely used air traffic corridors. Airspace occupancy risk is modeled using historical ADS-B–based flight trajectories, helicopter routes, and no-fly zones. A cost function derived from three-dimensional heatmaps discourages flight within or near heavily utilized airspace; these heatmaps are generated using either air traffic density or proximity to flight corridors and prohibited areas. Benchmarking results compare the risk-avoidance performance and computational efficiency of the search-based method with the minimum-risk three-dimensional Dubins solution under both airspace-only and joint airspace–ground risk minimization. The results demonstrate the potential of real-time emergency trajectory generation to significantly reduce exposure to high-risk airspace, thereby enhancing emergency landing safety and contributing to a safer, less disrupted airspace system in contingency. Future work will incorporate real-time traffic into a dynamic airspace model to capture time-varying operational constraints.

\section*{Acknowledgment}
The authors thank the OpenSky Network for providing air traffic data.

\bibliography{references}

\end{document}